\documentclass[10pt,twocolumn,letterpaper]{article}

\usepackage{float}
\usepackage[pagenumbers]{cvpr} 

\newcommand{\ours}{ICM}

\definecolor{cvprblue}{rgb}{0.21,0.49,0.74}
\usepackage[pagebackref,breaklinks,colorlinks,allcolors=cvprblue]{hyperref}

\def\paperID{} 
\def\confName{CVPR}
\def\confYear{2026}

\title{Attention, May I Have Your Decision? \\Localizing Generative Choices in Diffusion Models}

\author{
Katarzyna Zaleska$^{1}$* \quad
Łukasz Popek$^{1}$ \quad
Monika Wysoczańska$^{2}$ \quad
Kamil Deja$^{1,3}$\\
$^{1}$Warsaw University of Technology \quad $^{2}$valeo.ai \quad $^{3}$IDEAS Research Institute  \\
}

\begin{document}
\maketitle

\begingroup
\renewcommand\thefootnote{}
\footnotetext{*Corresponding author: \texttt{katarzyna.zaleska.me@gmail.com}}
\endgroup

\begin{abstract}
Text-to-image diffusion models exhibit remarkable generative capabilities, yet their internal operations remain opaque, particularly when handling prompts that are not fully descriptive. In such scenarios, models must make implicit decisions to generate details not explicitly specified in the text. This work investigates the hypothesis that this decision-making process is not diffuse but is computationally localized within the model's architecture. 
While existing localization techniques focus on prompt-related interventions, we notice that such explicit conditioning may differ from implicit decisions. Therefore, we introduce a probing-based localization technique to identify the layers with the highest attribute separability for concepts. Our findings indicate that the resolution of ambiguous concepts is governed principally by self-attention layers, identifying them as the most effective point for intervention. Based on this discovery, we propose \ours~ (Implicit Choice-Modification) -- a precise steering method that applies targeted interventions to a small subset of layers. Extensive experiments confirm that intervening on these specific self-attention layers yields superior debiasing performance compared to existing state-of-the-art methods, minimizing artifacts common to less precise approaches. The code is available at \url{https://github.com/kzaleskaa/icm}.
\end{abstract}
    
\section{Introduction}
\label{sec:intro}

\begin{figure}[h]
  \centering
  \resizebox{0.99\linewidth}{!}{\includegraphics{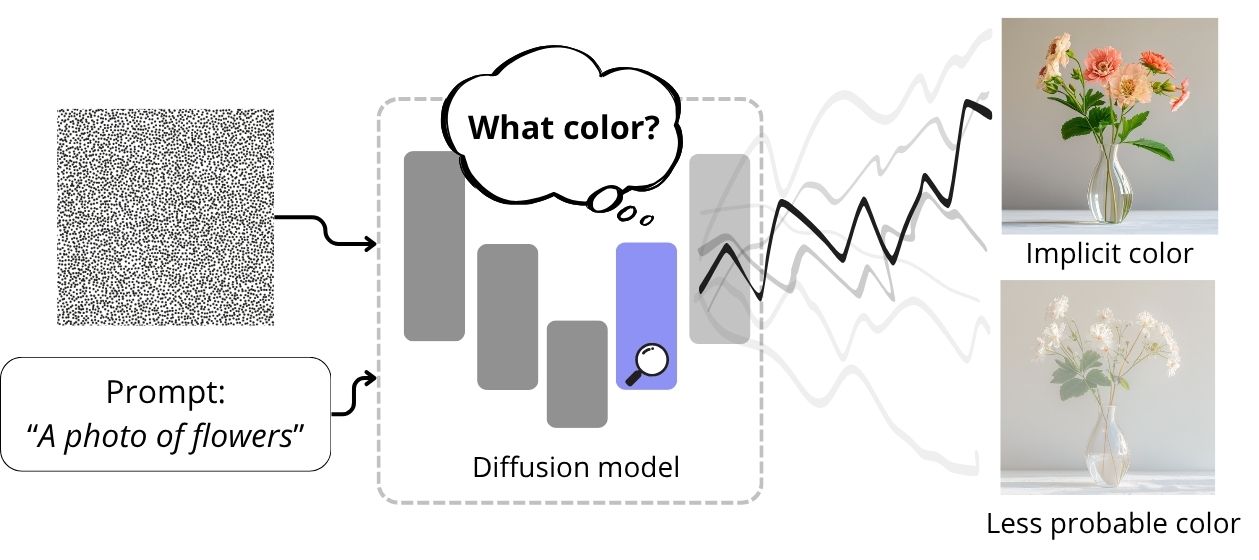}}
  \caption{
  We use linear probes to localize layers with the highest attribute separability for concepts not present in the prompt (e.g., a color when prompting for ``\emph{a photo of flowers}''). We show that we can use the learned probes to steer towards less probable outcomes.}
  \label{fig:teaser}
\end{figure}

Text-to-image (T2I) diffusion models~\citep{rombach2022highresolutionimagesynthesislatent} represent a major advance in AI, yet their internal mechanisms remain largely opaque, turning the whole process into a black-box. This opacity becomes particularly evident when text prompts lack specificity, forcing the model to make generative choices -- filling in missing details of the intended generation based on patterns learned from training data.
Often, these implicit decisions are benign; for instance, when prompted with \textit{``a photo of flowers"} the model must infer a color, shape, and background from the random noise to generate an image. However, the same underlying mechanism can lead to more problematic outcomes. These range from perpetuating social biases, such as defaulting to men for professional roles, to representational skews where a concept like 'USA President' consistently resolves to a single person, like Donald Trump. In this work, we want to understand where and how a diffusion model \textit{decides} on how to instantiate a general idea as presented in Figure~\ref{fig:teaser}.

The most common strategy for localizing knowledge in T2I models~\cite{basu2024-localizing-knowledge,basu2024mechanistic,staniszewski2025precise,park2025crossattention} is based on the idea of prompt injection, where a different text prompt is injected only in the selected layers to measure their impact on the output. Such precise localization unlocks a wide range of applications starting from direct concept editing or removal~\citep{gandikota2024unified,orgad2023editing}, through precise finetuning \cite{shirkavand2025efficient,everaert2023diffusion}, reduction in computational resources~\citep{liu2024faster} up to prevention of harmful content generation~\citep{staniszewski2025precise}. However, those prior works focus mostly on tracing concepts explicitly mentioned in the prompt, showing that their generation can be logically tied to the cross-attention layers responsible for integrating text and image representations. While reasonable for the analysis of objects or style localization, injecting attributes into prompts masks the internal mechanism the model uses when the prompt does not provide an answer.

In this work, we posit that the mechanism for implicit decision choices is computationally localized and distinct from the mechanism used for explicit text conditioning. To validate this, we introduce a probing-based localization technique that avoids the confounding effects of prompt engineering. Instead of injecting attributes into the input text, we generate images using underspecified prompts (e.g., 'a photo of a person') and retrospectively label the output attributes using an external classifier. Given such pseudo-labels, we train linear probes on the intermediate activations to quantify layer-wise discriminability. This allows us to pinpoint exactly where the model’s internal representation becomes linearly separable, identifying the layer at which the implicit decision about the selected attribute is most pronounced. 

Our localization technique reveals an unexpected finding. We notice that activations right after self-attention layers enable more precise linear separation between different instantiations of the same concept than activations after cross-attention layers, which are known to handle explicit prompts. This suggests that self-attention layers are responsible for discriminating between possible implicit options. To demonstrate the utility of our layer selection method, we apply it to debiasing in diffusion models. We build upon prior work that has addressed this task primarily through finetuning~\cite{shen2023finetuning,he2024debiasing} or activation steering~\cite{li2023self0discovering,parihar2025balancingactdistributionguideddebiasing,shi2025dissectingmitigatingdiffusionbias}, but propose to employ them exclusively to selected subset of layers for better precision. By intervening only in a subset of chosen layers,
 we achieve the state-of-the-art steering performance. We demonstrate that we can mitigate biases in gender, age, and race while minimizing the artifacts and quality degradation common in broader, less targeted interventions. Finally, we examine our localization technique on larger diffusion models with different architectures: the UNet-based SDXL~\cite{podell2023sdxl} and the Transformer-based SANA~\cite{xie2024sanaefficienthighresolutionimage}. Our contributions can be summarized as follows:
\begin{itemize}
    \item We show that with linear probing, we can localize the most important layers for changing implicit decisions in text-to-image diffusion models
    \item We highlight the important role self-attention layers play in the instantiation of general prompts into specific images, exhibiting higher attribute separability compared to cross-attention layers.
    \item We show that by focusing on the selected layers, we can improve the performance of debiasing with steering or fine-tuning.
\end{itemize}

\section{Related work}

In this section, we discuss literature related to our work across three main areas: understanding the decision-making process during generation (Section~\ref{sec:rw_2}), layer specialization and localization in diffusion models (Section~\ref{sec:rw_1}), and debiasing approaches for generative models as a practical application of aforementioned techniques (Section~\ref{sec:related_bias}).

\subsection{Decision process in Diffusion models}
\label{sec:rw_2}

In this work, we analyze how diffusion models make implicit decisions when prompts underspecify certain attributes, requiring the model to infer specific instantiations from initial noise. It has been demonstrated, however, that this initial noise significantly impacts generation outcomes~\cite{guo2024initno0}. 
From a text prompting perspective, several works have investigated prompt-to-image relationships, with~\cite{kong2023interpretable} establishing links between specific prompt elements and their visual manifestation in generated images.  Magnet~\cite{zhuang2024magnet} and Cat/Dog~\cite{chen2024cat} study the issues observed in explicit bindings between prompts and the outputs, while methods like Attend-and-Excite~\cite{chefer2023attend} or~\cite{li2024get} intervene in the cross-attention maps to mitigate the neglect of explicitly prompted subjects. Similarly, decisions on the semantic objects provided by cross-attention layers are used in the case of Prompt-to-Prompt~\cite{hertz2022prompt} and Ledits++~\cite{brack2024ledits++} to perform a precise edition.
Finally,~\citet{liu2024faster} shows that information is introduced into generated images through cross-attention layers, especially during initial denoising steps.

\subsection{Layers specialization in diffusion models}
\label{sec:rw_1}
Our work is related to a growing body of work examining how different types of \emph{knowledge} are distributed across layers in diffusion models. For example, Kwon et al.~\cite{kwon2023diffusionmodelssemanticlatent} show that the UNet's bottleneck can provide semantic representations for the diffusion models. Furthermore, works by ~\citet{basu2024-localizing-knowledge,basu2024mechanistic, liu2024towards} examining UNets, demonstrated that cross-attention layers are responsible for incorporating compositional and semantic information from the prompt, while self-attention modules focus more on the spatial layout. Similar localization patterns are also observed in the transformer architecture~\cite{zarei2025localizing}. 

A common approach for identifying layer specialization is \emph{activation patching} (causal tracing)~\cite{meng2022locating}, which establishes causal links by running the model on both \emph{clean} and \emph{corrupted} prompts, and swapping internal activations to determine which components are sufficient to alter the output. ~\citet{gandikota2024unified} builds on this idea by showing that applying closed-form adaptations across all cross-attention layers enables concept editing, debiasing, and removal of undesired content. Similarly, \citet{orgad2023editing} directly edits cross-attention weights using a closed-form solution to better align the source prompt with a target image.

\definecolor{mypurple}{RGB}{142, 146, 244}
\begin{figure*}[ht!]
  \includegraphics[width=\textwidth, trim={5 0 0 5}, clip]{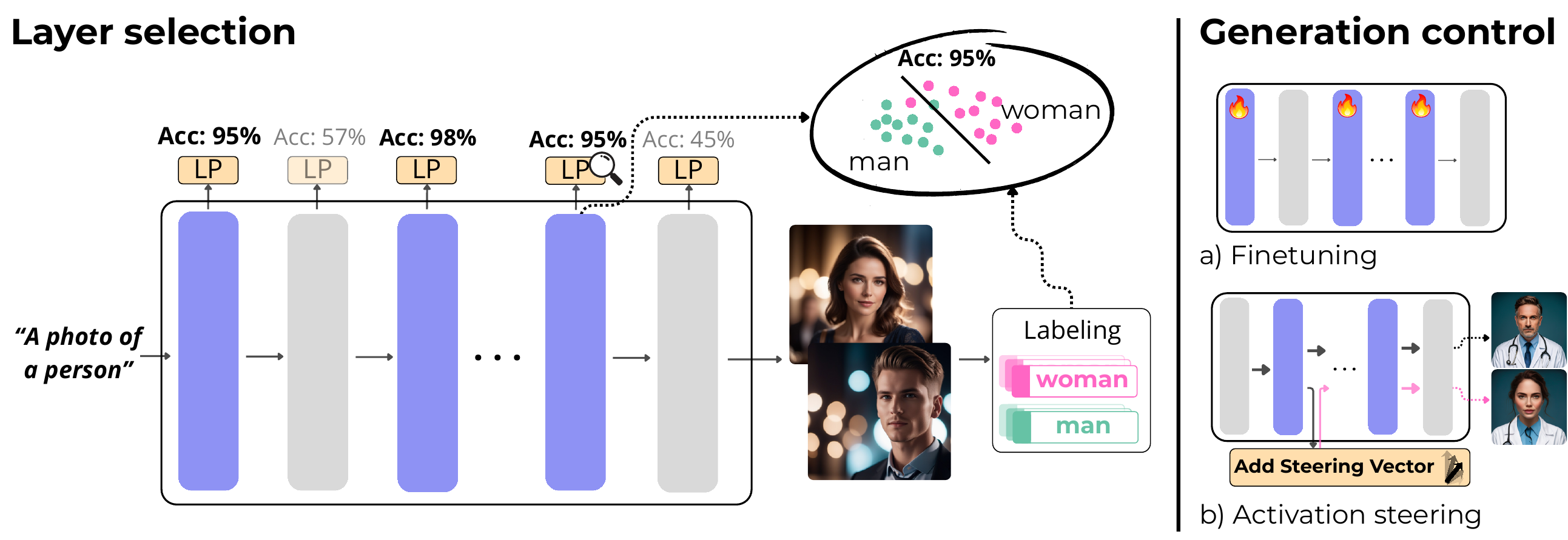}
  \caption{\textbf{Overview of \ours.} We identify optimal layers for steering by measuring their discriminability using an external classifier. Layers are ranked by the classification accuracy of a linear probing (denoted here as LP) on their activations, and the top-performing layers (here in \textcolor{mypurple}{purple}) are selected for targeted intervention. The selected layers can then be used for two applications: generation control via finetuning (top right) or activation steering (bottom right), enabling precise attribute manipulation while preserving image quality.}
  \label{fig:main_method}
\end{figure*}

\subsection{Debiasing in generative models}
\label{sec:related_bias}
Recent research on debiasing T2I diffusion models has explored a variety of strategies to mitigate stereotypical associations and fairness issues. These approaches can be broadly categorized into two groups based on \emph{finetuning} or \emph{activation steering}.

In the first group, \citet{shen2023finetuning} frames the problem as a distribution mismatch, introducing a new loss used for finetuning that enforces the target distribution. \citet{he2024debiasing} further explores this idea with an iterative distribution alignment procedure by minimizing the KL-divergence of the target distributions.

 Activation steering is an alternative approach that removes the need for finetuning through inference-time intervention. For example, \cite{li2023self0discovering} introduces a method that automatically discovers latent directions in the UNet bottleneck, which can be used to steer generations towards a randomly selected target. Similarly, \citet{parihar2025balancingactdistributionguideddebiasing} uses guidance from a classifier trained on the same h-space of the UNet bottleneck, to guide towards unbiased generations. To enable more fine-grained steering in the UNet's bottleneck, \citet{shi2025dissectingmitigatingdiffusionbias} employs sparse autoencoders trained on SD mid-block activations to identify features responsible for specific attribute realizations. These features are then used to steer generation following the approach of~\cite{li2023self0discovering}.

While the aforementioned works focus on debiasing pretrained T2I models, \citet{kim2024training} explores how to prevent biases from emerging during the initial training phase.

A notable limitation shared by the works described above is their approach to layer selection. While different methods experiment with various finetuning objectives or search for optimal steering directions, they ultimately apply modifications either to all cross-attention layers or to heuristically chosen components like the UNet mid-block.
In this work, we argue that, similarly to what has been observed in Large Language Models (LLMs)~\citep{limisiewicz2023debiasing,qin2025lftf0,chandna2025dissecting,meng2022locating} those original techniques can be further improved with a precise localization of the layers responsible for introducing biases to the generations.

\section{Method}

In this section, we present our approach for selective layer manipulation in diffusion models -- \ours~ (Implicit Choice-Modification). To achieve precise control, we first identify which layers represents best separability of specific choices for the general concept through activation-based classification (Section~\ref{sec:localization}). Then, we demonstrate how to control the generation of these choices by applying targeted interventions (fine-tuning or activation steering) exclusively to the selected layers (Section~\ref{sec:control}).

\subsection{Linear Probing for Layer Selection}
\label{sec:localization}
We aim at selecting a subset of layers which yields
highest concept separability for the general prompt rather than localizing such layers on the basis of explicit prompt injections. We propose to do it through linear probing. Such an approach allows us to consider all of the model layers beyond the ones directly conditioned on text. The process (visualized in Figure~\ref{fig:main_method}) is comprised of three stages: (1) generating images and collecting intermediate activations using general prompts, (2) post-hoc pseudo-labeling of generated samples via an external classifier, and (3) training linear probes to quantify layer-wise concept discriminability.

 \paragraph{Activation Extraction.}

 We define a general concept $\mathcal{C}$ which a model can instantiate into one of several mutually exclusive options $\mathcal{A} = \{a_1, a_2, \dots, a_K\}$. For example, if $\mathcal{C}$ represents \textit{Gender} the set of instantiations might be $\mathcal{A} = \{\text{male}, \text{female}\}$. To analyze the model's intrinsic bias and decision-making process, we construct a general prompt $p_{gen}$ that describes $\mathcal{C}$ without any specification $a_k \in \mathcal{A}$ (e.g., $p_{gen} = \textit{``a photo of a person"}$).

We generate a dataset of $N$ images, $\mathcal{X} = \{x^{(i)}\}_{i=1}^N$, by conditioning the model solely on $p_{gen}$. During the generation of each sample $x^{(i)}$, we systematically extract the intermediate internal activations. Let $H_{l,t}^{(i)} \in \mathbb{R}^{d_l}$ represent the activation vector, which we obtain through average pooling of layer's activations $l \in \mathcal{L}$ and denoising timestep $t \in \mathcal{T}$ for the $i$-th sample, where $\mathcal{L}$ is the set of all analyzed layers and $d_l$ is the feature dimension at layer $l$.

\paragraph{Pseudo-Labeling via External Classification.}
To recover implicit decisions of the model, for each generated image $x^{(i)}$, we obtain a label $y^{(i)}$:$$y^{(i)} = \Phi(x^{(i)}), \quad \text{where } y^{(i)} \in \mathcal{A}$$
We utilize an external classifier (e.g., CLIP-based), creating a labeled dataset of internal representations $\mathcal{D}_{l,t} = \{(H_{l,t}^{(i)}, y^{(i)})\}_{i=1}^N$. This approach ensures that we analyze the activations corresponding to the model's actual choices rather than forcing choices via prompt engineering.

\paragraph{Localization via Linear Probing.}

 Finally, to identify the location of the most important layers, we measure how linearly separable the attributes in $\mathcal{A}$ are within the activation space of each layer. For every layer $l$ and timestep $t$, we fit a logistic regression-based probe $f_{l,t}: \mathbb{R}^{d_l} \rightarrow \mathcal{A}$.

The probe is fit to predict the attribute $y^{(i)}$ given the activation $H_{l,t}^{(i)}$. The discriminability of layer $l$ at timestep $t$ is quantified by the accuracy of $f_{l,t}$ on a training set. Layers yielding high classification accuracy indicate a strong 
link to the final visual outcome concept $\mathcal{C}$.

\begin{figure}[t]
        \centering
        \includegraphics[width=\linewidth]{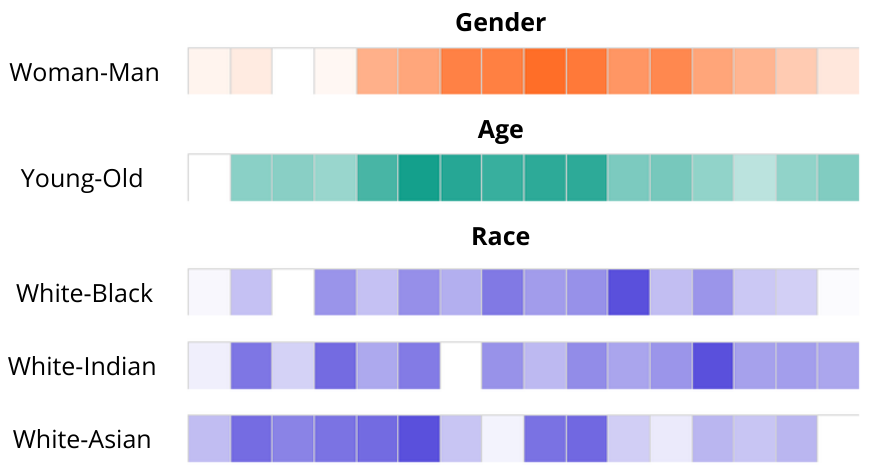}
        \vspace{-1.5em}
        \label{fig:percentage_injected_vs_none_sd}
\caption{Comparison of mean accuracy across all timesteps for the examined concepts, evaluated on the test set.}
    \label{fig:percentage_injected_vs_none_all}
\end{figure}

\subsection{Generation Control} 
\label{sec:control}

Having identified the layers that are most sensitive to specific concept attributes through our classification-based selection procedure, we now discuss how these selected layers can be leveraged to control the generation process.

\paragraph{Activation steering.}
To evaluate our localization and enable inference-time control over model's implicit decisions, we exploit the classifiers from Section~\ref{sec:localization} to form an activation steering vectors. For each selected layer $\ell$ and timestep $t$, the trained logistic regression classifier $f_{\ell}$ provides a weight vector $\hat{w}_{\ell,t}$ that defines a linear decision boundary in the activation space. We normalize this weight vector to obtain the steering direction:
\[
s_{\ell,t} = \frac{\hat{w}_{\ell,t}}{\|\hat{w}_{\ell,t}\|_2}.
\]
This unit vector $s_{\ell,t}$ captures the axis of maximum class separability and serves as the optimal direction for manipulating the implicit decision. At generation time, we modify the forward pass by steering the activations at the selected layers according to:
\[
H_{\ell,t}' = H_{\ell,t} + \alpha \, s_{\ell,t},
\]
where $H_{\ell,t}$ represents the unmodified activation tensor and the scaling factor $\alpha$ controls the magnitude of the intervention, with larger absolute values producing stronger attribute modifications.

\paragraph{Fine-tuning.} 
Finally, we can benefit from our localization method when we want to fine-tune the model by adapting only the weights of the selected layers. More specifically, 
for a selected decision that we want to reinforce,
we generate $K$ images using class-specific prompts. 
We construct the dataset as triplets $(x, p_{gen}, p_{spec})$, where $x$ denotes an image generated from a specific prompt $p_{spec}$ 
but reported during finetuning under the corresponding general prompt $p_{gen}$. 
For example, if $p_{spec}= \textit{``a face of a young doctor''}$ and $p_{gen} = \textit{``a face of a doctor''}$, 
then $x$ is generated with $p_{spec}$ and paired with $p_{gen}$ for training. We then train a model with the Low-Rank Adaptation (LoRA) technique  \citep{hu2021loralowrankadaptationlarge} using the standard diffusion loss, which minimizes the mean squared error between the true noise $\epsilon$ and the model prediction $\hat{\epsilon}_\theta$ at timestep $t$ for a given noised sample $x_t$ and conditioning prompt $p$:

\begin{equation}
\mathcal{L} = \mathbb{E}_{\epsilon, t}\Big[ \lVert \epsilon - \hat{\epsilon}_{\theta}(x_t, p, t) \rVert_2^2 \Big]
\end{equation}

\begin{table*}[h]
  \centering
    \caption{Results for gender, age, and race debiasing with SD v1.5. Our \ours methods achieve a competitive Fairness Discrepancy (FD $\downarrow$) while best preserving image quality (FID $\downarrow$) and prompt-text alignment (CLIP-T $\uparrow$). Results of competing methods from \cite{shi2025dissectingmitigatingdiffusionbias}.} \vspace{-1em}
  \label{tab:debias_results}
  \resizebox{0.98\linewidth}{!}{
  \begin{tabular}{@{}lcccccccccccc@{}}
  \toprule
  \multirow{2}{*}{\textbf{Method}} & \multicolumn{4}{c}{\textbf{Gender (2)}} & \multicolumn{4}{c}{\textbf{Age (3)}} & \multicolumn{4}{c}{\textbf{Race (4)}} \\
  \cmidrule(lr){2-5}\cmidrule(lr){6-9}\cmidrule(lr){10-13}
   & FD $\downarrow$ & FID $\downarrow$ & CLIP-I $\uparrow$ & CLIP-T $\uparrow$
   & FD $\downarrow$ & FID $\downarrow$ & CLIP-I $\uparrow$ & CLIP-T $\uparrow$
   & FD $\downarrow$ & FID $\downarrow$ & CLIP-I $\uparrow$ & CLIP-T $\uparrow$ \\
  \midrule
  Original               & 0.564 & 120.06 & -      & 0.6155 & 0.752 & 120.06 & -      & 0.6155 & 0.558 & 120.06 & -      & 0.6155 \\
  \midrule
    Finetuning   ~\cite{shen2024finetuningtexttoimagediffusionmodels}    & \underline{0.050} & 161.47 & \underline{0.8779} & 0.6095 & 0.746 & 161.47 & {0.8779} & 0.6095 & \underline{0.198} & 161.47 & 0.8779 & 0.6095 \\
  \textbf{\ours~(Finetuning)}& 0.535 & 143.98 & \textbf{0.9187} & \textbf{0.6189} & 0.681 & 122.56 & \underline{0.9075} & \textbf{0.6198} & 0.449 & 123.47 & 0.9020 & \textbf{0.6188} \\
  \midrule
  Latent Editing   \cite{kwon2023diffusionmodelssemanticlatent} & 0.408 & 166.11 & 0.8253 & 0.6005 & 0.682 & 200.90 & 0.8527 & {0.6122} & 0.524 & 153.05 & 0.8804 & 0.6086 \\
  H-Distribution ~\cite{parihar2025balancingactdistributionguideddebiasing} & 0.222 & 151.68 & 0.8475 & 0.6087 & 0.506 & 147.71 & 0.8345 & 0.6098 & 0.544 & 126.90 & 0.8255 & 0.6100 \\
  Latent Direction  ~\citep{li2024selfdiscoveringinterpretablediffusionlatent} & 0.305 & 129.37 & 0.8058 & {0.6091} & \underline{0.052} & \underline{113.81} & 0.8151 & 0.6067 & \textbf{0.175} & 128.30 & 0.8211 & 0.6132 \\
  
  DIFFLENS ~\cite{shi2025dissectingmitigatingdiffusionbias}& \textbf{0.046} & \textbf{112.83} & 0.8501 & 0.6090 & \textbf{0.049} & \textbf{99.17} & 0.8778 & 0.6057 & 0.401 & \underline{119.86} & \underline{0.9096} & {0.6149} \\

\textbf{ \ours~(Steering)} & 0.087 & \underline{122.08} & 0.8500 & \underline{0.6140} & 0.133 & 114.87 & \textbf{0.9195} &  \underline{0.6150} & 0.266 & \textbf{116.88} & \textbf{0.9099} & \underline{0.6172}  \\
  \bottomrule
  \end{tabular}}
\vspace{-1em}
\end{table*}

\section{Experiments}
In this section, we demonstrate how our method can be effectively applied to the practical use case of debiasing diffusion models. We start by introducing an experimental setup, including models and evaluation metrics we use to validate our approach. Then, we present our main results, in which we employ the proposed localization technique in a practical scenario for removing societal biases such as Gender, Age, and Race from the model. In addition to comparing with recent state-of-the-art approaches, we present a thorough experimental study of various aspects of the proposed solution. Finally, we scale our experiments to more advanced models with different architectures.

\subsection{Experimental setup}

\textbf{Models.} We evaluate the decision process across different diffusion architectures, including (1) U-Net based models -- Stable Diffusion (SD) and Stable Diffusion XL (SDXL) \citep{podell2023sdxl}, which rely on CLIP-like encoders for text embeddings, and (2) transformer-based SANA model \citep{xie2024sanaefficienthighresolutionimage}, which uses large language model (LLM)-based encoders to generate text embeddings. 

\begin{figure}[t]
    \centering

    \captionsetup{skip=6pt}
    \begin{minipage}{0.23\textwidth}
        \centering
        \includegraphics[width=\linewidth]{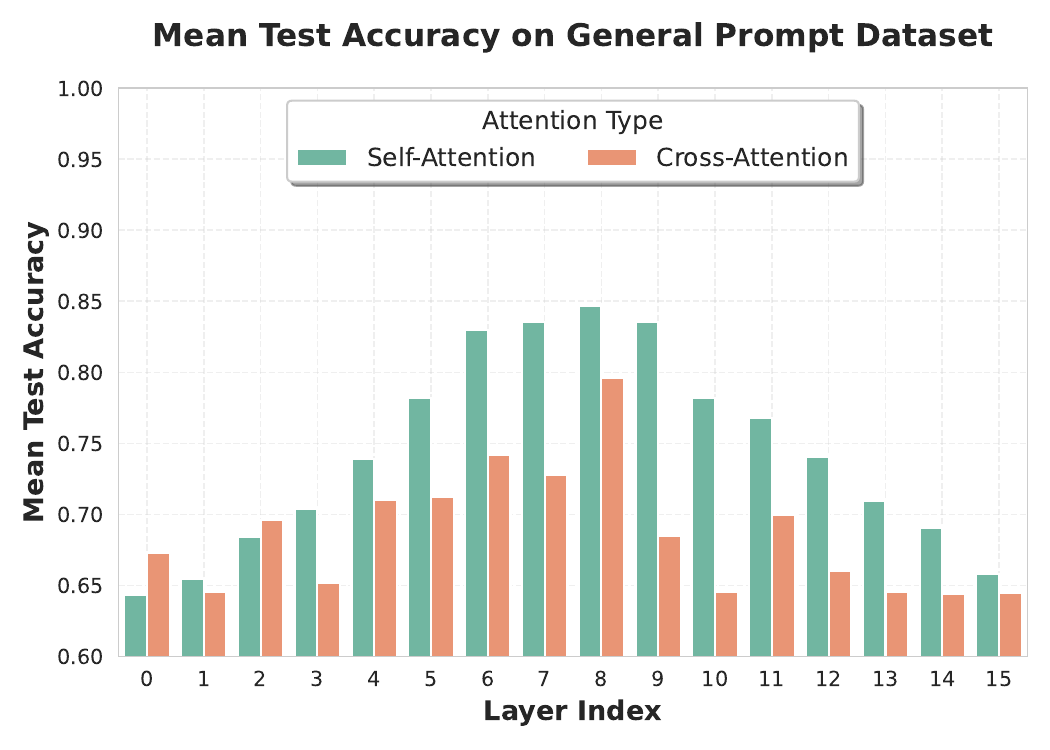}
        \subcaption{General prompts}
        \label{fig:general_prompts}
    \end{minipage}\hfill
    \begin{minipage}{0.23\textwidth}
        \centering
        \includegraphics[width=\linewidth]{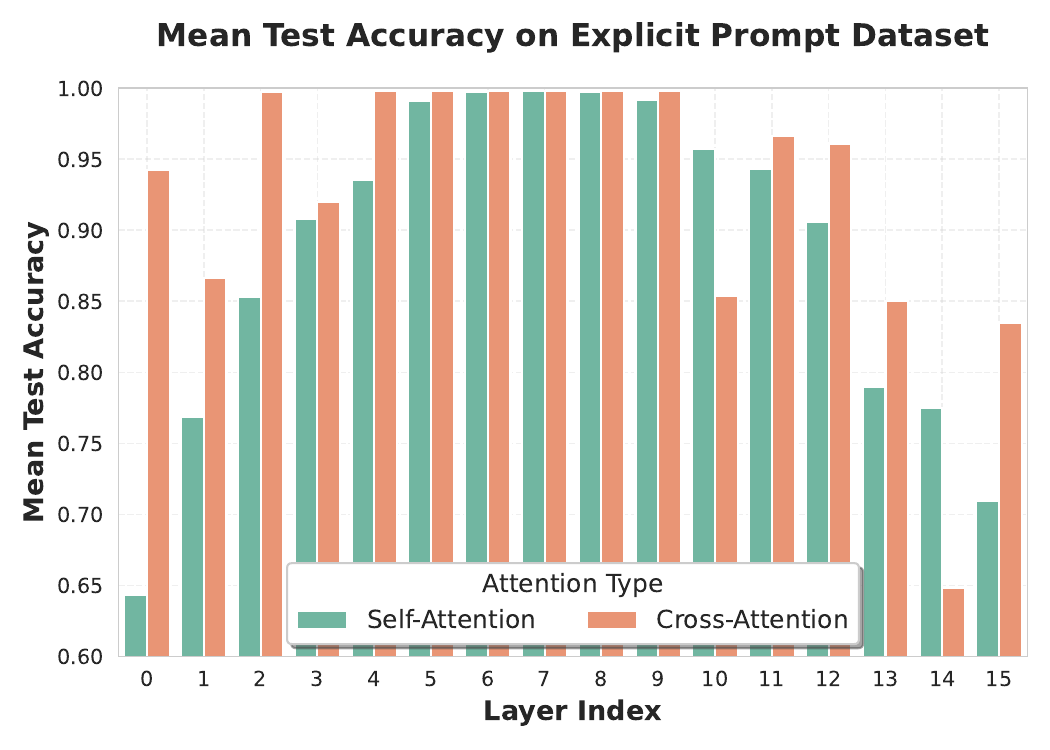}
        \subcaption{Explicit prompts}
        \label{fig:specific_prompts}
    \end{minipage}

    \caption{Accuracies of linear probes trained to predict gender from 
    activations collected after self- and cross-attention layers of the 
    SD~1.5 UNet. Self-attention layers exhibit generally higher 
    discriminative power than cross-attention layers, with a clear peak 
    towards the middle block. Results are compared for the general 
    prompts~\subref{fig:general_prompts} and explicit 
    prompts~\subref{fig:specific_prompts}.}
    \label{fig:probe_cross_vs_self}
\end{figure}

\paragraph{Evaluation metrics.}
Following \citet{shi2025dissectingmitigatingdiffusionbias}, we measure bias using Fairness Discrepancy (FD) \cite{parihar2025balancingactdistributionguideddebiasing} as:
\begin{equation}
\mathrm{FD} = \left\lVert \bar{p} - \mathbb{E}_{\mathbf{x} \sim p_\theta(\mathbf{x})} (\mathbf{y}) \right\rVert_2
\end{equation}
where $\bar{p}$ denotes the reference distribution over attributes, 
$\mathbf{x} \sim p_\theta(\mathbf{x})$ represents samples drawn from the model, 
and $\mathbf{y}$ is the predicted attribute distribution of these samples. 
We generate 500 images per prompt and evaluate gender (male, female), age (young: 0--19, adult: 20--59, old: 60+), and race (white, black, Asian, Indian) using FairFace \citep{kärkkäinen2019fairfacefaceattributedataset}.
Additionally, we measure image quality via FID against the FFHQ~\cite{karras2019stylebasedgeneratorarchitecturegenerative} using 2000 generated images. We also use CLIP-I to measure similarity to reference images and CLIP-T to measure alignment with input text prompts. We discuss the employed metrics in more detail in Section~\ref{supp:metrics} of Supplementary Material.

\paragraph{Implementation details.} We evaluate ~\ours~on Stable Diffusion v1.5~\cite{stable_diffusion} as the main debiasing model. We start the steering mechanism from timestep $t=15$ when the performance of the linear probes significantly exceeds a random guess. We select between 2 and 4 self-attention layers that stand out in terms of performance. Given the significant differences between tasks and bias severity, for each scenario, we individually select the $\alpha$ steering strength. For some cases, where the model was not able to generate enough samples from the target class given the general prompt (e.g., old people in age debiasing), for localization, we extend a set of general prompts to specify the target (e.g., \textit{``a person with gray hair"}). We give more details of the implementation in Section \ref{app:exp_setup} of Supplementary Material.

\begin{table*}[t]
\centering
\caption{Debiasing performance comparison when applying steering after Self-Attention vs. Cross-Attention layers. Self-Attention yields a dramatically better Fairness Discrepancy (FD), while maintaining comparable image quality (FID) and prompt alignment (CLIP-T).\vspace{-1em}}
\label{tab:self_vs_cross}
\resizebox{\textwidth}{!}{
\begin{tabular}{@{}lcccccccccccc@{}}
\toprule
& \multicolumn{4}{c}{\textbf{Gender (2)}} & \multicolumn{4}{c}{\textbf{Age (3)}} & \multicolumn{4}{c}{\textbf{Race (4)}} \\
\cmidrule(r){2-5} \cmidrule(r){6-9} \cmidrule(r){10-13}
\textbf{Method} & \textbf{FD ↓} & \textbf{FID ↓} & \textbf{CLIP-I ↑} & \textbf{CLIP-T ↑} & \textbf{FD ↓} & \textbf{FID ↓} & \textbf{CLIP-I ↑} & \textbf{CLIP-T ↑} & \textbf{FD ↓} & \textbf{FID ↓} & \textbf{CLIP-I ↑} & \textbf{CLIP-T ↑} \\
\midrule
Steer cross-attn & 0.365 & 118.39 & \textbf{0.8631} & \textbf{0.6171} & 0.612 & 114.26 & \textbf{0.8991} & \textbf{0.6167} & 0.428 & \textbf{120.28} & \textbf{0.8561} & \textbf{0.6179} \\
Steer self-attn & \textbf{0.085} & \textbf{118.31} & {0.8556} & 0.6136 & \textbf{0.273} & \textbf{112.57} & 0.8954 & {0.6153} & \textbf{0.298} & {129.51} & {0.8403} & {0.6172} \\
\midrule
Finetuning cross-attn & 0.535 & 143.98 & 0.9187 & \textbf{0.6189} & \textbf{0.681} & \textbf{122.56} & 0.9075 & \textbf{0.6198} & \textbf{0.449} & \textbf{123.47} & 0.9020 & \textbf{0.6188} \\
Finetuning self-attn & \textbf{0.463} & \textbf{139.04} & \textbf{0.9408} & 0.6175 & 0.770 & 131.25 & \textbf{0.9280} & 0.6186 & 0.524 & 124.36 & \textbf{0.9148} & 0.6186 \\
\bottomrule
\end{tabular}
}

\end{table*}

\subsection{Main results}

To validate the impact of our localization technique on the precision of debiasing we compare in Table~\ref{tab:debias_results} our approach against several recent state-of-the-art debiasing methods, namely: Latent Editing \cite{kwon2023diffusionmodelssemanticlatent},  H-Distribution \cite{parihar2025balancingactdistributionguideddebiasing}, Latent Direction \cite{li2024selfdiscoveringinterpretablediffusionlatent}, Finetuning \cite{shen2024finetuningtexttoimagediffusionmodels}, and DIFFLENS \cite{shi2025dissectingmitigatingdiffusionbias}. A significant challenge for this task is the common trade-off between reducing bias and maintaining the model's overall generative quality and alignment with the text prompt. The results in Table~\ref{tab:debias_results} highlight this trade-off. For instance, while methods like Finetuning or Latent Editing can reduce Fairness Discrepancy, they do so at the expense of the image quality, as shown by their high FID scores and prompt-text alignment. 
Overall, our \ours (Steering) variant provides the best balance. It achieves a strong reduction in bias (0.087 FD for Gender) while maintaining great image quality (122.08 FID) and superior prompt alignment. 

\subsection{Self or cross-attention?}
\label{sec:self-cross}

As mentioned in Section~\ref{sec:rw_2}, prior work has shown that cross-attention layers primarily encode semantic information from text prompts, while self-attention modules handle spatial layout~\cite{liu2024towards}. Consequently, most interventions target cross-attention layers in the U-Net bottleneck, assuming biases are semantic concept-level problems. While this is reasonable for explicitly prompted concepts, in our context of underspecified prompts (\textit{``a photo of a person"}), cross-attention remains agnostic as it lacks explicit attribute tokens. This implies that the decision must originate from the initial noise. We hypothesize that self-attention layers act as the deciding mechanism by propagating and solidifying independent stochastic cues (e.g., hairstyle in one part of the image with the presence of lipstick in another) into a unified generation (and hence, deciding on the gender).

We therefore compare probes trained on activations right after the self-attention layers with the ones trained after the cross-attention. We observe that probes trained after self-attention are substantially more accurate at classifying the implicit decisions. As presented in Figure~\ref{fig:probe_cross_vs_self}, probes in both of the cases work best around the middle block, but in general, the performance of the self-attention probes is significantly higher.
To further evaluate the implications of those discrepancies, we compare the performance of probing-based steering performed on the same transformer blocks, but either after self- or cross-attention layers. As shown in Table~\ref{tab:self_vs_cross}, steering self-attention layers is significantly more effective for debiasing (as measured by the FD metric) without significantly influencing the image quality and alignment metrics. Our results suggest that implicit decisions are not semantic choices governed by cross-attention, but are instead translated from the initial random noise by self-attention, making it the true locus of these decisions and the most effective point for intervention.

\subsection{Layer selection -- visual comparison}
\begin{figure}[t]
    \includegraphics[width=1.0\linewidth]{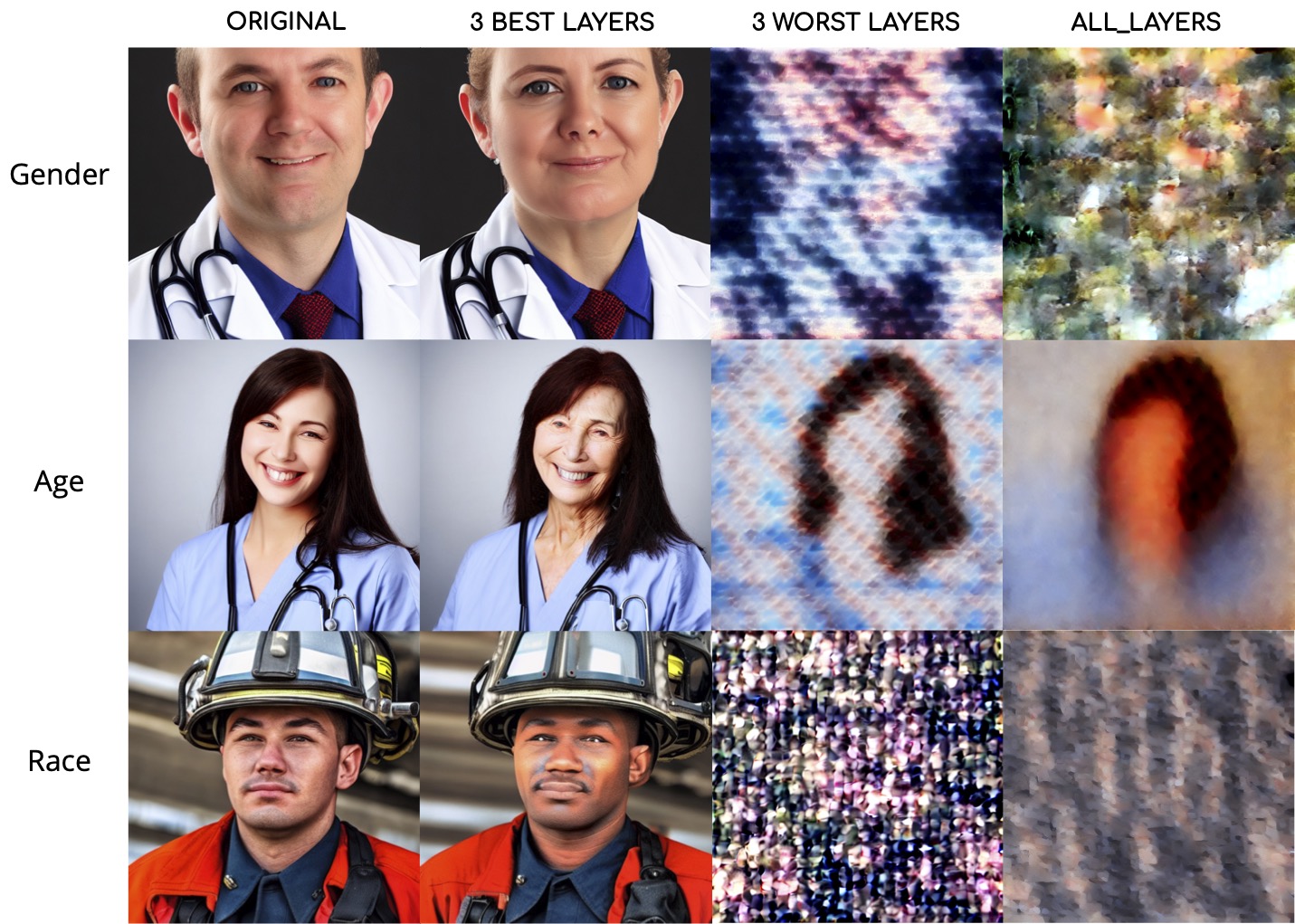}
    \caption{\textbf{Visual ablation of layer selection} for activation steering. \ours selects the 3 best layers to preserve image quality while achieving desired modifications, compared to steering: 1)~the 3 worst-performing layers, or 2) all layers without selection.
    }
    \label{fig:layer_ablation}
\end{figure}

We present in Figure~\ref{fig:layer_ablation} a visual comparison of steering effects across different layer selections in the Stable Diffusion v1.5. We show (from left to right): original generated images, results when applying steering to the 3 best-performing layers, results from the 3 worst-performing layers, and results from steering all layers simultaneously. The comparison demonstrates that strategic layer selection is critical for effective steering—targeting optimal layers preserves image quality and achieves desired modifications (column 2), while poorly chosen layers introduce severe artifacts and degradation (column 3). Steering all layers indiscriminately also leads to quality deterioration, highlighting the importance of selective layer manipulation for maintaining generation fidelity.

\begin{figure*}[t]
  \centering
\includegraphics[width=0.9\linewidth]{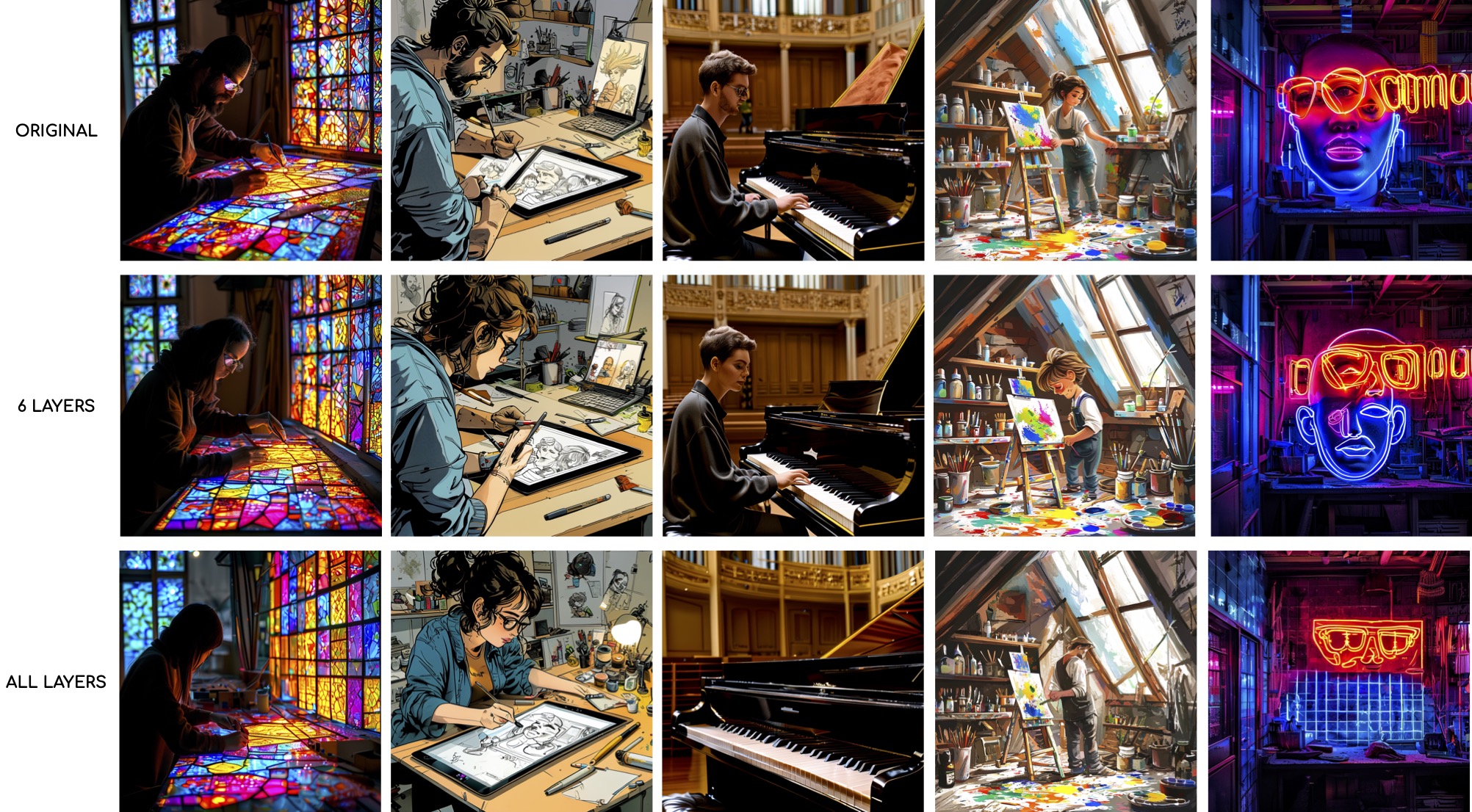}
  \caption{Example generations from the SANA model with applied gender steering. We compare steering using only 6 best performing probes on self-attention layers with steering using all of the layers.\vspace{-1em}}
  \label{fig:sana_gender}
\end{figure*}

\subsection{Linear Probing or Prompt Injection?}

As discussed in Section~\ref{sec:rw_2}, prompt injection can effectively identify layers that influence the output towards a specific attribute. It relies, however, on introducing an explicit conditioning (e.g. \textit{``a photo of a man"}) into a generation process initiated by a generic prompt. In this work, we posit a hypothesis that such external steering does not necessarily reflect the model's intrinsic mechanisms for making a decision.
To validate this hypothesis, we analyze whether a representational gap exists between implicit decisions and explicit conditioning. We train linear probes to distinguish between activations corresponding to naturally occurring implicit choices versus those forced by explicit prompts. These generations are subsequently classified by an external CLIP model to identify instances where the model implicitly decided (e.g., on \textit{man} or a \textit{woman}). We then compare these implicit activations against those collected from generations using specific prompts (e.g. \textit{``a photo of a man"}).

As presented in Table~\ref{tab:probe_accuracies_generic_specific} the results of the probe analysis confirm our hypothesis. Linear probes trained to differentiate between the two generation modes achieve test accuracies significantly higher than 50\% chance baseline. Such performance indicates a distribution shift between the content generated via implicit choices versus explicit prompting. This finding suggests that the internal mechanisms driving default decision-making are distinct from those involved during explicit conditioning.

\begin{table}[t]
\small
\centering
\caption{Train and test accuracies for linear probes distinguishing between activations from generic prompt (\textit{``a photo of a person"}) and explicit (specific prompt) generation. }
\label{tab:probe_accuracies_generic_specific}
\begin{tabular}{@{} l c c@{}}
\toprule
 \textbf{Specific Prompt} & {\textbf{Train Acc.}} & {\textbf{Test Acc.}} \\
\midrule
 \textit{``a photo of a man"} & 70.37 & 62.08 \\
 \textit{``a photo of a woman"} & 73.37 & 70.54 \\
\addlinespace 
\textit{``a photo of an older person"} & 84.24 & 68.63 \\
\textit{``a photo of a younger person"} & 82.85 & 88.86\\
\addlinespace 
\textit{``a photo of a dark hair person"} & 92.84 & 88.63 \\
\textit{``a photo of a light hair person"} & 88.79 & 80.40\\
\addlinespace 
\textit{``a photo of a happy person"} & 80.38 & 79.30 \\
\textit{``a photo of a sad person"} & 73.69 & 56.77\\
\bottomrule
\end{tabular}
\end{table}
 Table~\ref{tab:specific_general} compares debiasing performance on the Gender task using steering with linear probes fit on general vs. specific prompts. We observe that general prompts which allow for more precise capture of the distribution modes lead to on par debiasing with a smaller effect on the general image quality and alignment.

\begin{table}[t]
  \centering
  \small
    \caption{Comparison of steering performance when linear probes are trained on activations from general vs. specific prompts (Gender task). The same settings (the $\alpha$ value and start timestep) and layers were used for each run.}
  \label{tab:specific_general}
  \begin{tabular}{@{}lcccc@{}}
  \toprule
   & FD $\downarrow$ & FID $\downarrow$ & CLIP-I $\uparrow$ & CLIP-T $\uparrow$\\
   \midrule
\textbf{General} & 0.089 & 119.45 & 0.883 & 0.613  \\
\textbf{Specific} &0.087 & 122.08 &0.850  &  0.614 \\
  \bottomrule
  \end{tabular}
\end{table}

\begin{figure*}[h!]
  \centering
\includegraphics[width=0.9\linewidth]{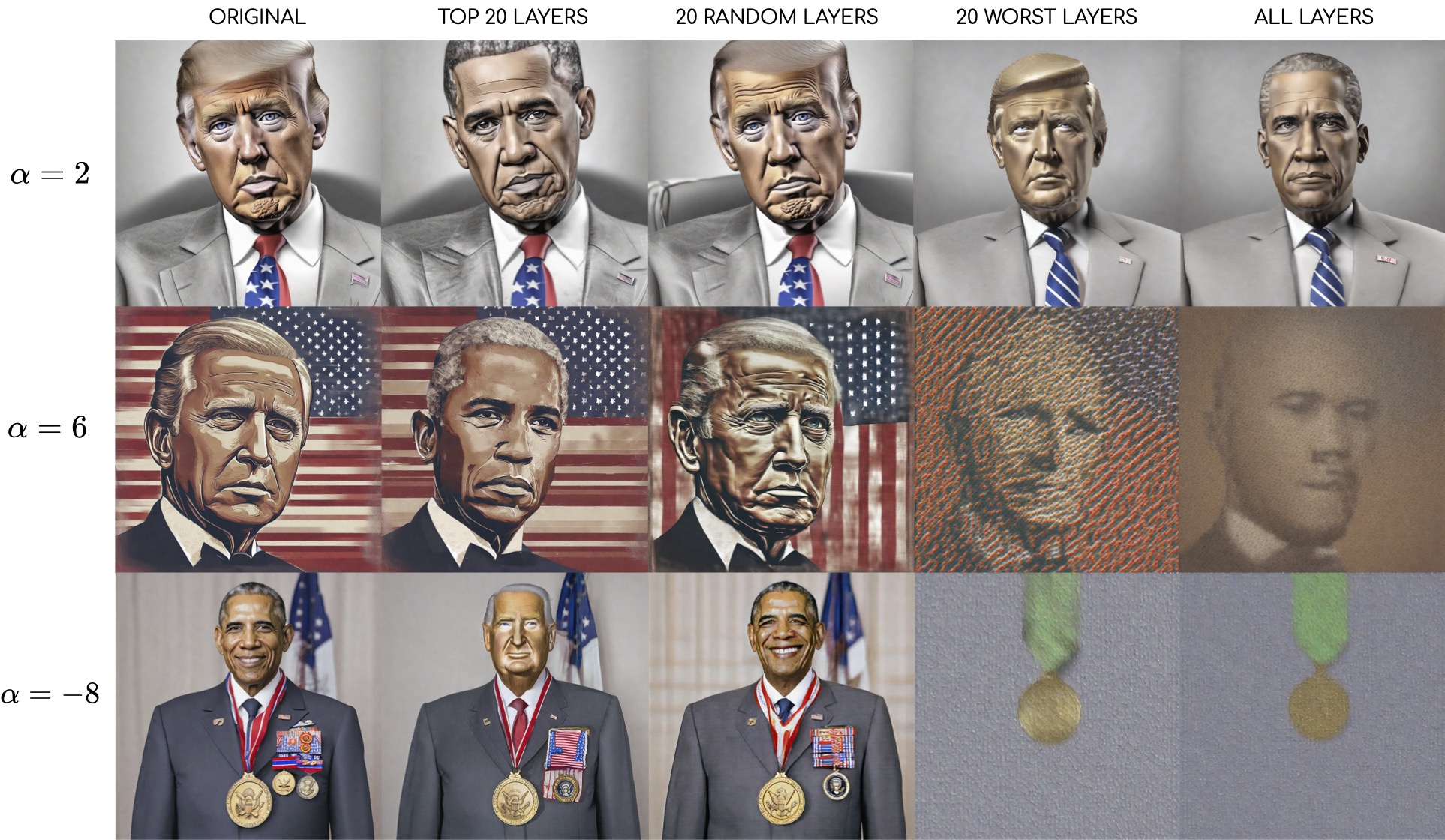}
  \caption{Example generations from SDXL model steered using probes trained on prompts describing a USA president, where positive class is `Barack Obama', and negative is all of the other generated presidents.\vspace{-1em}}
  \label{fig:sdxl_presidents}
\end{figure*}

\subsection{Controlled generation with large models}

Finally, we also scale our experiments beyond the SD v1.5 for larger models. We consider SDXL~\cite{podell2023sdxl} a model with 70 self- and cross-attention layers, and SANA~\cite{xie2024sanaefficienthighresolutionimage}, which is a diffusion transformer with 20 transformer blocks (20 self and cross attention layers).

For SANA, we once more relate to the problem of societal biases, localizing the layers that decide on gender. We collect activations from all of the self-attention layers using a general prompt. We then split them into male/female classes and train linear probes to select 6 out of 20 layers with the highest accuracy. As presented in Figure~\ref{fig:sana_gender}, interventions with the selected subset of layers are more precise, resulting in smaller differences in the background when compared to steering all of the layers.

For SDXL, we extend our analysis beyond societal biases by examining the prompt "a photo of the USA president," which typically produces images of Donald Trump, Barack Obama, or Joe Biden. Our probing technique identifies 20 of 70 self-attention layers as critical for this decision. Figure~\ref{fig:sdxl_presidents} shows steering results toward "Barack Obama" using: selected high-discriminability layers, random layers, low-discriminability layers, and all layers. Steering with only the selected layers better preserves the original generation and is more robust to varying steering strengths. At higher scaling factors $\alpha$
(including negative values in the bottom row), steering with poorly selected or all layers severely degrades image quality.
To quantify the trade-off between steering effectiveness and image preservation, we steered 100 generations of \textit{``a photo of the USA president"} towards \textit{Barack Obama}. As shown in Table~\ref{tab:sdxl_steering}, steering all layers is highly effective, changing 91.0\% of images, but severely degrades the image, as indicated by the low CLIP-I (0.779). In contrast, steering 20 random layers preserves the original image (0.932 similarity) but is ineffective, only successfully steering 51.0\% of the images. Our method, which targets only the top 20 localized layers, provides the best balance: it achieves a strong steering success rate (83.0\%) while maintaining high image fidelity and achieving the highest average classifier confidence.

\begin{table}[t]
\centering
\caption{Comparison of steering 100 images towards \textit{Barack Obama} using different layer selection strategies. Our method (Top 20) achieves effective steering while maintaining the highest image similarity to the original generations, avoiding the severe image degradation seen when steering all layers.}

\label{tab:sdxl_steering}
  \resizebox{0.98\linewidth}{!}{
\begin{tabular}{@{}lccc@{}}
\toprule
\textbf{Layer Selection} & \textbf{Steering Success (\%)} & \textbf{Avg. Confidence} & \textbf{CLIP-I $\uparrow$} \\
\midrule
\textbf{Top 20 (Ours)} & 83.0\% & 0.917 & 0.893 \\
Random 20 & 51.0\% & 0.838 & 0.932 \\
All Layers & 91.0\% & 0.879 & 0.779 \\
\bottomrule
\end{tabular}
}
\vspace{-1em}
\end{table}

\section{Conclusions}
This work investigates the internal mechanisms governing implicit decision-making in text-to-image diffusion models. Through probing-based localization, we demonstrate that the resolution of ambiguous prompts is principally governed by self-attention layers. Leveraging this insight, we introduce \ours{}, a method for precise intervention on these specific layers. Our experiments confirm that \ours{} achieves superior debiasing performance compared to state-of-the-art methods, effectively mitigating gender, age, and race biases while preserving image fidelity. In addition to our main results, we provide a thorough evaluation of the rationale behind the main design choices and show that our approach scales to large diffusion models across different architectures.

\section*{Acknowledgments}

We thank Łukasz Staniszewski for all the help and insightful feedback during this project. This work was funded by the National Science Centre, Poland, grant no UMO-2023/51/B/ST6/03004. The computing resources were provided by the PL-Grid Infrastructure, grant no.: PLG/2025/018390 and PLG/2025/018424. This paper has been supported by the Horizon Europe Programme (HORIZON-CL4-2022-HUMAN-02) under the project "ELIAS: European Lighthouse of AI for Sustainability", GA no. 101120237.
    \small
    \bibliographystyle{ieeenat_fullname}
    \bibliography{main}

\clearpage
\setcounter{page}{1}
\maketitlesupplementary

\definecolor{softgray}{gray}{0.95}
\tcbset{
  mypromptbox/.style={
    enhanced,
    colback=softgray,      
    colframe=white,        
    boxrule=0pt,           
    arc=2pt,               
    left=6pt, right=6pt, top=4pt, bottom=4pt,  
    fontupper=\sffamily\small, 
    width=\linewidth
  }
}

\appendix

\section{Implementation details on Steering vectors}
\begin{figure}[h]
    \centering
    \includegraphics[width=\linewidth]{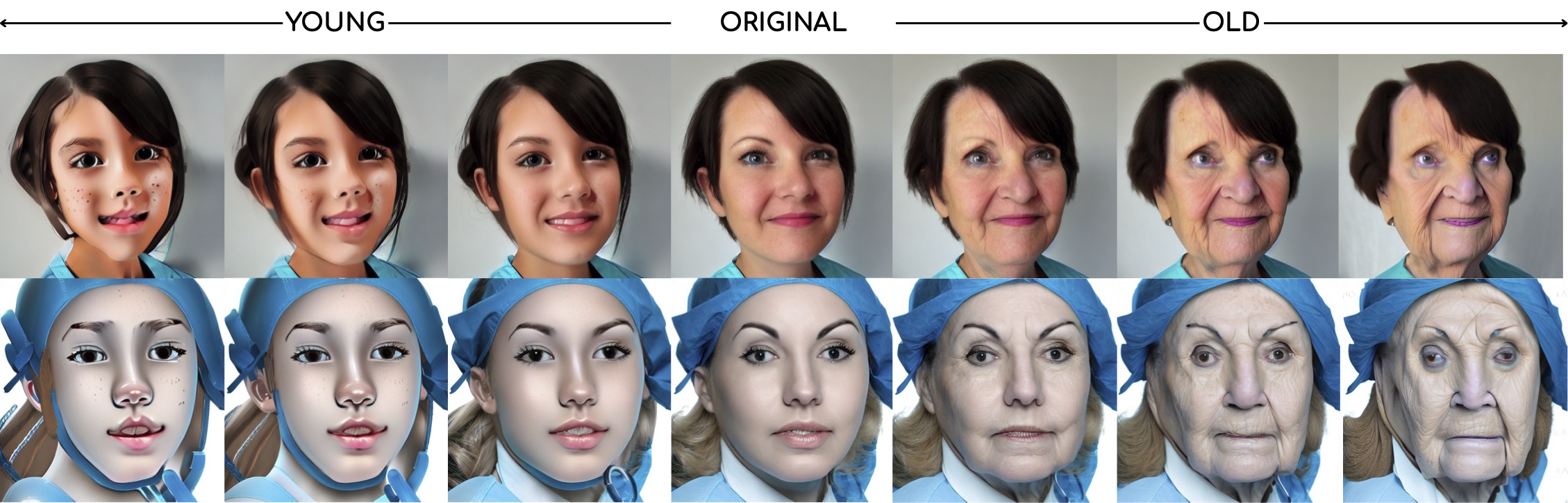}
    \caption{Effect of increasing $\alpha$ values along the young–old direction. Larger $\alpha$ produces stronger age-related changes.}
    \label{fig:young_old_alpha}
\end{figure}

To calculate the steering vectors discussed in Section~\ref{sec:control}, we use a pipeline composed of a StandardScaler\footnote{\url{scikit-learn.org/stable/modules/generated/sklearn.preprocessing.StandardScaler.html}} and a LogisticRegression\footnote{\url{scikit-learn.org/stable/modules/generated/sklearn.linear_model.LogisticRegression.html}} model
(with a maximum of 1000 iterations) on the pooled activations for each layer–timestep pair. After training, we map the learned coefficients back to the original, unstandardized feature space by dividing the weight vector by the corresponding scaling factors. The rescaled vector is then normalized to unit length:
\[
s_{\ell,t} = \frac{\hat{w}_{\ell,t}}{\|\hat{w}_{\ell,t}\|_2}.
\]
yielding the final steering vector \( s_{\ell,t} \) for layer \(\ell\) and timestep \(t\). The vector is used later during generation as: 
\[
H_{\ell,t}' = H_{\ell,t} + \alpha \, s_{\ell,t},
\]
where $H_{\ell,t}$ denotes the unmodified activation tensor and the scaling factor $\alpha$ controls the magnitude of the intervention. We design the probes in a binary fashion. For example, if the vector is trained with \textit{old} as the positive class, positive $\alpha$ values shift the generation toward older appearances, whereas negative $\alpha$ values shift it toward not-old ones. The magnitude of $\alpha$ determines the strength of the modification, as illustrated in Figure~\ref{fig:young_old_alpha}. In the backward diffusion pass with classifier-free guidance, the steering vector is applied only to the component conditioned on the text prompt.

\section{Additional details on experimental Setup}
\label{app:exp_setup}

For the steering-based debiasing, we introduce a random component that selects the direction in which the entire batch of images is shifted. When there are $n$ possible decisions, each direction is chosen with probability $\frac{1}{n}$. For example, for race we consider four options - white, black, asian, and indian - so each has a probability of $0.25$.

We use steering vectors trained as binary classifiers for the following pairs: woman–man, young–old, white–black, white–indian, and white–asian. Since most generated images are classified as adult or white, we do not apply modifications for the adult or white directions. For binary directions (e.g., age), the steering vector supports both positive and negative $\alpha$ values, enabling movement toward either side of the decision boundary. All selected layers and their corresponding $\alpha$ values are summarized in Table~\ref{tab:steering_params}.

\begin{table}[h]
    \centering
    \caption{Parameters used for steering across different directions.\vspace{-1em}}
    \begin{tabular}{l p{5.5cm} c}
        \toprule
        \textbf{Direction} & \textbf{Layers} & \textbf{\(\alpha\)} \\
        \midrule

        \multirow{4}{*}{woman}
            & up\_blocks.1.attn.2.t\_blocks.0.attn1 & \multirow{4}{*}{-10} \\
            & up\_blocks.1.attn.1.t\_blocks.0.attn1 & \\
            & up\_blocks.1.attn.0.t\_blocks.0.attn1 & \\
            & mid\_block.attn.0.t\_blocks.0.attn1   & \\ \hline

        \multirow{4}{*}{man}
            & up\_blocks.1.attn.2.t\_blocks.0.attn1 & \multirow{4}{*}{10} \\
            & up\_blocks.1.attn.1.t\_blocks.0.attn1 & \\
            & up\_blocks.1.attn.0.t\_blocks.0.attn1 & \\
            & mid\_block.attn.0.t\_blocks.0.attn1   & \\  \hline\hline

        \multirow{3}{*}{old} 
            & up\_blocks.1.attn.1.t\_blocks.0.attn1 & \multirow{3}{*}{8}  \\ 
            & mid\_block.attn.0.t\_blocks.0.attn1 &  \\
            & up\_blocks.1.attn.0.t\_blocks.0.attn1 &  \\ \hline

        \multirow{3}{*}{young} 
            & up\_blocks.1.attn.1.t\_blocks.0.attn1 & \multirow{3}{*}{-8}  \\ 
            & mid\_block.attn.0.t\_blocks.0.attn1 &  \\
            & up\_blocks.1.attn.0.t\_blocks.0.attn1 &  \\ \hline

        adult & -- & -- \\ \hline \hline

        \multirow{2}{*}{black}
            & up\_blocks.1.attn.1.t\_blocks.0.attn1 & \multirow{2}{*}{15} \\
            & mid\_block.attn.0.t\_blocks.0.attn1 &  \\ \hline

        \multirow{3}{*}{indian}
            & mid\_block.attn.0.t\_blocks.0.attn1 & \multirow{3}{*}{10} \\
            & up\_blocks.1.attn.2.t\_blocks.0.attn1 &  \\
            & down\_blocks.2.attn.0.t\_blocks.0.attn1 &  \\ \hline

        \multirow{3}{*}{asian}
            & up\_blocks.1.attn.1.t\_blocks.0.attn1 & \multirow{3}{*}{10} \\
            & mid\_block.attn.0.t\_blocks.0.attn1 &  \\
            & up\_blocks.1.attn.0.t\_blocks.0.attn1 &  \\ \hline
        
        white & -- & -- \\
        
        \bottomrule
    \end{tabular}
    \label{tab:steering_params}
\end{table}

\begin{figure*}[t]
    \centering
    \includegraphics[width=\textwidth]{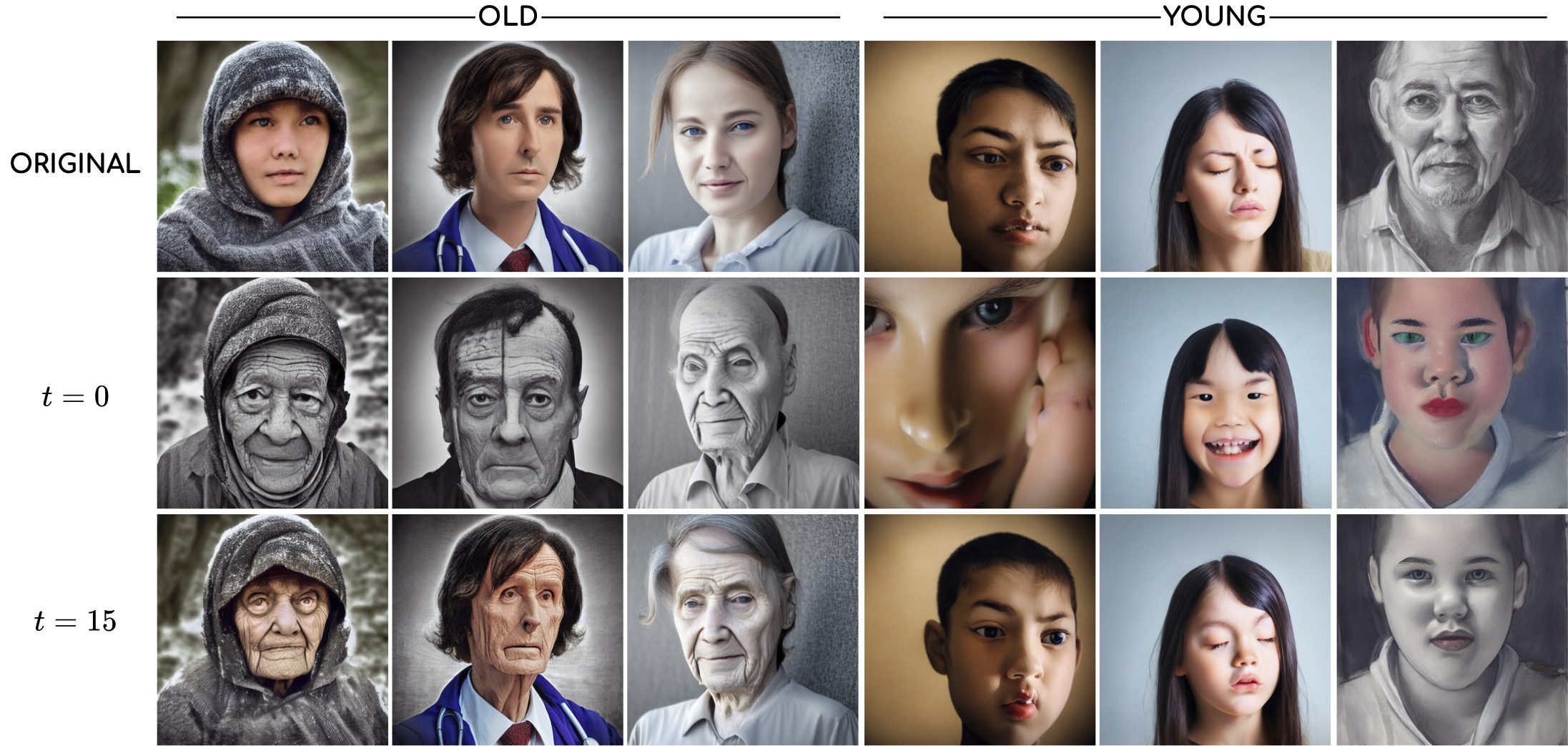}
    \caption{Example generations showing that applying the steering vector at later timesteps preserves the overall image structure while still shifting the predicted age direction. Here, $t$ denotes the timestep at which the modification begins.}
    \label{fig:appendix_timesteps}
\end{figure*}

\begin{figure}[t]
    \centering
    \includegraphics[width=0.98\linewidth]{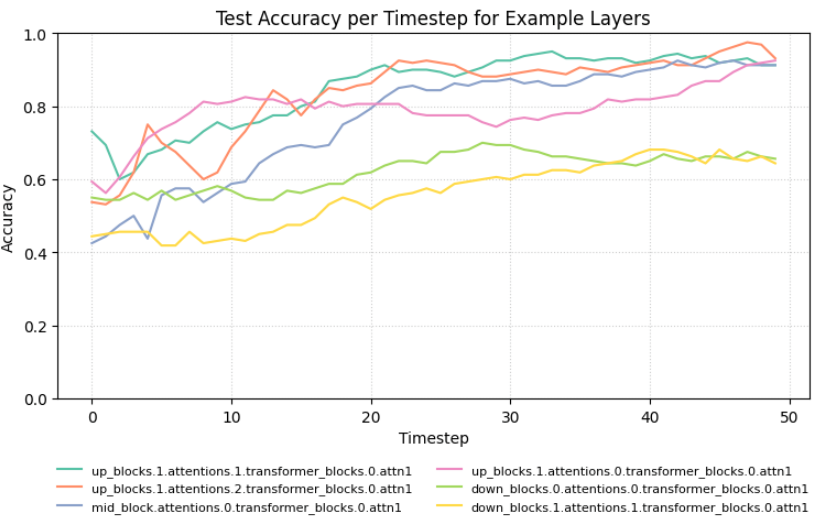}
    \caption{Test accuracy across the selected layers and timesteps for gender dataset.}
    \label{fig:timestep_accuracy}
\end{figure}

We apply each modification only after the first 15 timesteps, as the logistic-regression classifiers exhibit lower accuracy during the initial stages of denoising. Figure~\ref{fig:timestep_accuracy} shows the test accuracy across several selected layers, illustrating that accuracy increases after the early timesteps. Applying the intervention later in the denoising process better preserves the structure of the generated images, as shown in Figure~\ref{fig:appendix_timesteps}.

\begin{table*}[h]
  \centering
  \caption{Finetuning comparison for cross-attention layers selected via prompt injection versus randomly chosen layers.\vspace{-1em}}
  \resizebox{0.98\linewidth}{!}{
  \begin{tabular}{@{}lcccccccccccc@{}}
  \toprule
  \multirow{2}{*}{\textbf{Method}} & \multicolumn{4}{c}{\textbf{Gender (2)}} & \multicolumn{4}{c}{\textbf{Age (3)}} & \multicolumn{4}{c}{\textbf{Race (4)}} \\
  \cmidrule(lr){2-5}\cmidrule(lr){6-9}\cmidrule(lr){10-13}
   & FD $\downarrow$ & FID $\downarrow$ & CLIP-I $\uparrow$ & CLIP-T $\uparrow$
   & FD $\downarrow$ & FID $\downarrow$ & CLIP-I $\uparrow$ & CLIP-T $\uparrow$
   & FD $\downarrow$ & FID $\downarrow$ & CLIP-I $\uparrow$ & CLIP-T $\uparrow$ \\
  \midrule
  Original               & 0.564 & 120.06 & -      & 0.6155 & 0.752 & 120.06 & -      & 0.6155 & 0.558 & 120.06 & -      & 0.6155 \\
  \midrule
Finetuning (rank=32, selected) & \textbf{0.515} & \textbf{128.27} & 0.9028 &  0.6172  & \textbf{0.699} & \textbf{114.34} & \textbf{0.8943} & 0.6173 &  \textbf{0.485} & 127.90 & 0.9062 & \textbf{0.6190} \\
Finetuning (rank=32, random) & 0.542 & 133.62 & \textbf{0.9230} & \textbf{0.6177}  & 0.733 & 117.61 & 0.8916 & \textbf{0.6191} & 0.537 & \textbf{123.64} & \textbf{0.9412} & 0.6159 \\
  \multicolumn{13}{@{}l@{}}{}\\[-1.2ex]
  \bottomrule
  \end{tabular}}
  \label{tab:appendix_debias_results}
\end{table*}

\section{Additional details on fine-tuning}
We finetune Stable Diffusion v1.5 using low-rank adaptation (LoRA), applying rank-32 (for gender concept in main results) and rank-64 (for age and race) adapters to selected layers while keeping all other parameters frozen. We optimize the model with AdamW using a learning rate of \(3\times10^{-5}\), a cosine learning rate schedule, and 1000 warmup steps. Training is performed with mixed-precision \texttt{bf16}, gradient clipping with a maximum norm of $1$, and gradient checkpointing. We resize images to a resolution of \(512\times512\) with center cropping and random horizontal flips. We use a batch size of 2 with 4 gradient accumulation steps. The model is trained for 30 epochs with 8 data-loader workers. To align with previous works, in \ours(Finetuning), we finetune the model using only images generated by the model itself, which directly impacts the performance of the finetuned model in terms of FID. 

For the main debiasing comparison, gender and age fine-tuning used the same layers as for steering; race used the union of all layers from the black, Indian, and Asian experiments (Table~\ref{tab:steering_params}).

We compare fine-tuning only a selected subset of layers identified by prompt injection (described in section \ref{appendix:prompt_injection_experiment_details}) against fine-tuning random layers. The results in Table~\ref{tab:appendix_debias_results} show that updating only the selected layers achieves stronger performance on most metrics. 

\section{Evaluation metrics}
\label{supp:metrics}

We measure bias using Fairness Discrepancy (FD) \citep{shi2025dissectingmitigatingdiffusionbias,parihar2025balancingactdistributionguideddebiasing}, which is the Euclidean distance between a reference and a generated distribution:

\begin{equation}
\mathrm{FD} = \left\lVert \bar{p} - \mathbb{E}_{\mathbf{x} \sim p_\theta(\mathbf{x})} (\mathbf{y}) \right\rVert_2
\end{equation}

where $\bar{p}$ is the reference distribution, $\mathbf{x}$ are model samples, and $\mathbf{y}$ is their predicted attribute distribution.
A lower FD value means the generated distribution is closer to the reference. For Stable Diffusion v1.5, we generate $500$ images per prompt, compute metrics for each, and report the average. FairFace \citep{kärkkäinen2019fairfacefaceattributedataset} provides predictions for gender, age, and race. Following prior work, we use two gender classes (male, female), three age groups (young: 0–19, adult: 20–59, old: 60+), and four race groups (white, black, Asian - combining East and Southeast Asian - and Indian).

We assess image quality using the Fréchet Inception Distance (FID), with the FFHQ \citep{karras2019stylebasedgeneratorarchitecturegenerative} dataset as the reference set of real images. FID measures the distance between generated and real image distributions, with lower values indicating closer alignment.

We also use CLIP-I to assess alignment with reference images and CLIP-T to assess alignment with the input prompts. For this, we generated original image embeddings $\mathbf{e}_{\text{img}}^{\text{orig}}$ and debiased image embedding $\mathbf{e}_{\text{img}}^{\text{gen}}$ to calculate CLIP-I metric:

\begin{equation}
\text{CLIP-I} = 
\frac{\mathbf{e}_{\text{img}}^{\text{orig}} \cdot \mathbf{e}_{\text{img}}^{\text{gen}}}
{\lVert \mathbf{e}_{\text{img}}^{\text{orig}} \rVert \, \lVert \mathbf{e}_{\text{img}}^{\text{gen}} \rVert}.
\end{equation}

For textual alignment, we use text prompt embedding $\mathbf{e}_{\text{text}}$ and debiased images $\mathbf{e}_{\text{img}}^{\text{gen}}$:

\begin{equation}
\text{CLIP-T} = 
\frac{\mathbf{e}_{\text{text}} \cdot \mathbf{e}_{\text{img}}^{\text{gen}}}
{\lVert \mathbf{e}_{\text{text}} \rVert \, \lVert \mathbf{e}_{\text{img}}^{\text{gen}} \rVert}.
\end{equation}

For embedding extraction, we use the CLIP ViT-L/14 model \citep{radford2021learningtransferablevisualmodels}. 

\section{Details on prompt templates}
For training logistic regression models, we have two prompt versions: (1) a general prompt without decision information, and (2) a specific prompt containing direct information.

Examples of general prompts:
\begin{tcolorbox}[mypromptbox]
a face of a person\\[0.4em]
photo of a person\\[0.4em]
portrait photo of a person\\[0.4em]
a close-up face of a person\\[0.4em]
a detailed portrait of a person\\[0.4em]
studio photo of a person\\[0.4em]
professional photo of a person\\[0.4em]
a headshot of a person\\[0.4em]
a professional studio headshot of a person\\[0.4em]
portrait of a person in natural light\\[0.4em]
portrait of a person indoors
\end{tcolorbox}

Examples of specific prompts for gender:
\begin{tcolorbox}[mypromptbox]
a portrait of a \textbf{woman} in natural light\\[0.4em]
photo of a \textbf{man} wearing casual clothes\\[0.4em]
a detailed portrait of a \textbf{lady} with soft lighting\\[0.4em]
a close-up face of a \textbf{gentleman} with calm expression\\[0.4em]
portrait of a \textbf{girl} outdoors in daylight\\[0.4em]
a headshot of a \textbf{woman} against a plain background\\[0.4em]
studio photo of a \textbf{man} with neutral expression\\[0.4em]
portrait of a \textbf{boy} with gentle expression\\[0.4em]
photo of a \textbf{woman} taken in warm sunset light\\[0.4em]
a professional photo of a \textbf{man} in front of a window\\[0.4em]
portrait of a \textbf{lady} with confident look
\end{tcolorbox}

\section{Additional qualitative results}
\begin{figure}[h]
    \centering
    \includegraphics[width=\linewidth]{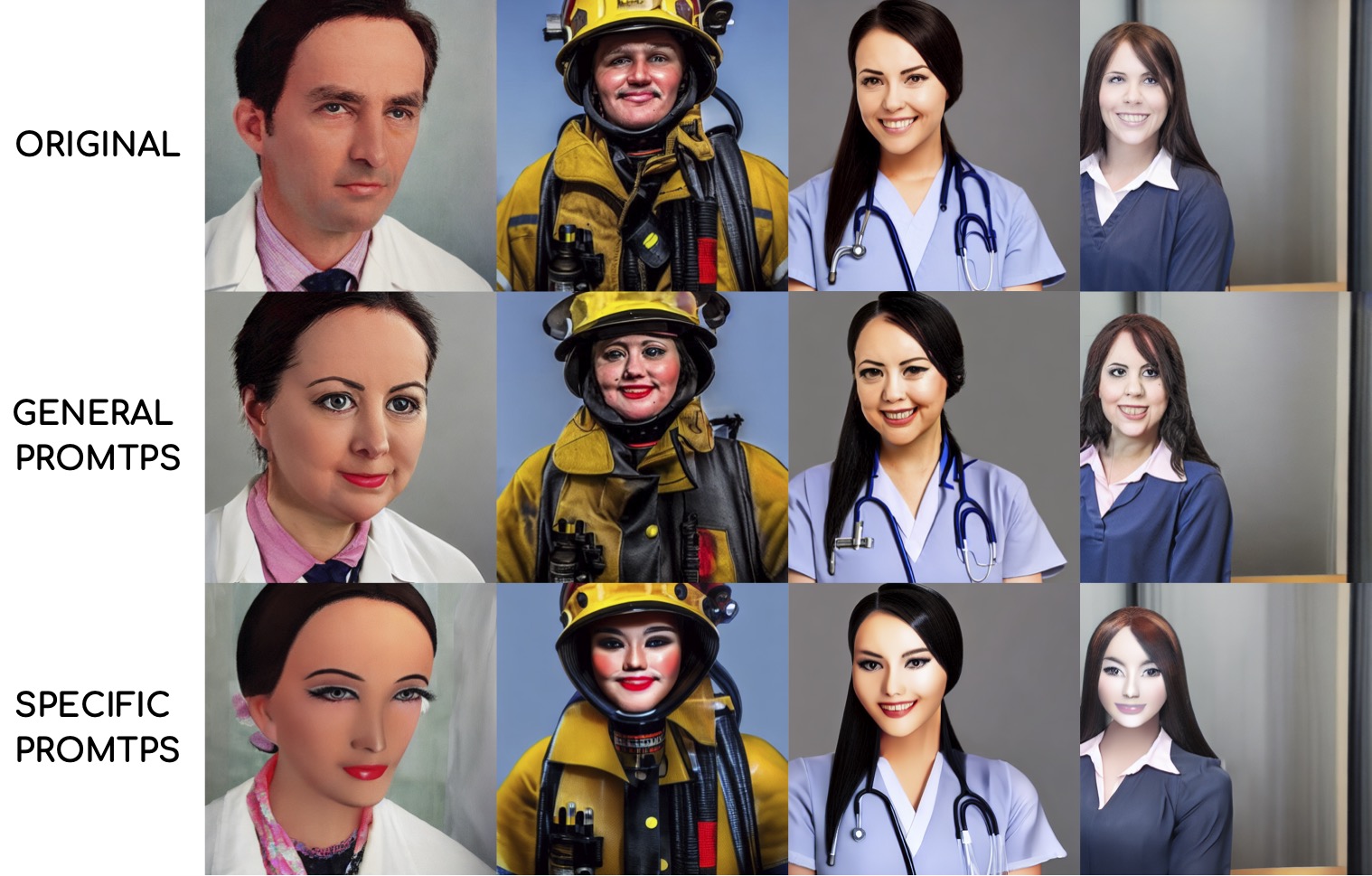}
    \caption{Comparison of generations from the original model (top), steering trained on general prompts (middle), and steering trained on specific prompts (bottom).}
    \label{fig:general_specific}
\end{figure}

We observe that training logistic regression classifiers on general prompts yields less artificial, more natural images than training on highly specific prompts. Figure~\ref{fig:general_specific} shows outputs from the original model and from steering vectors trained on general or specific prompts, all using the same~$\alpha$.

\begin{figure*}[t]
    \centering
    \includegraphics[width=\linewidth]{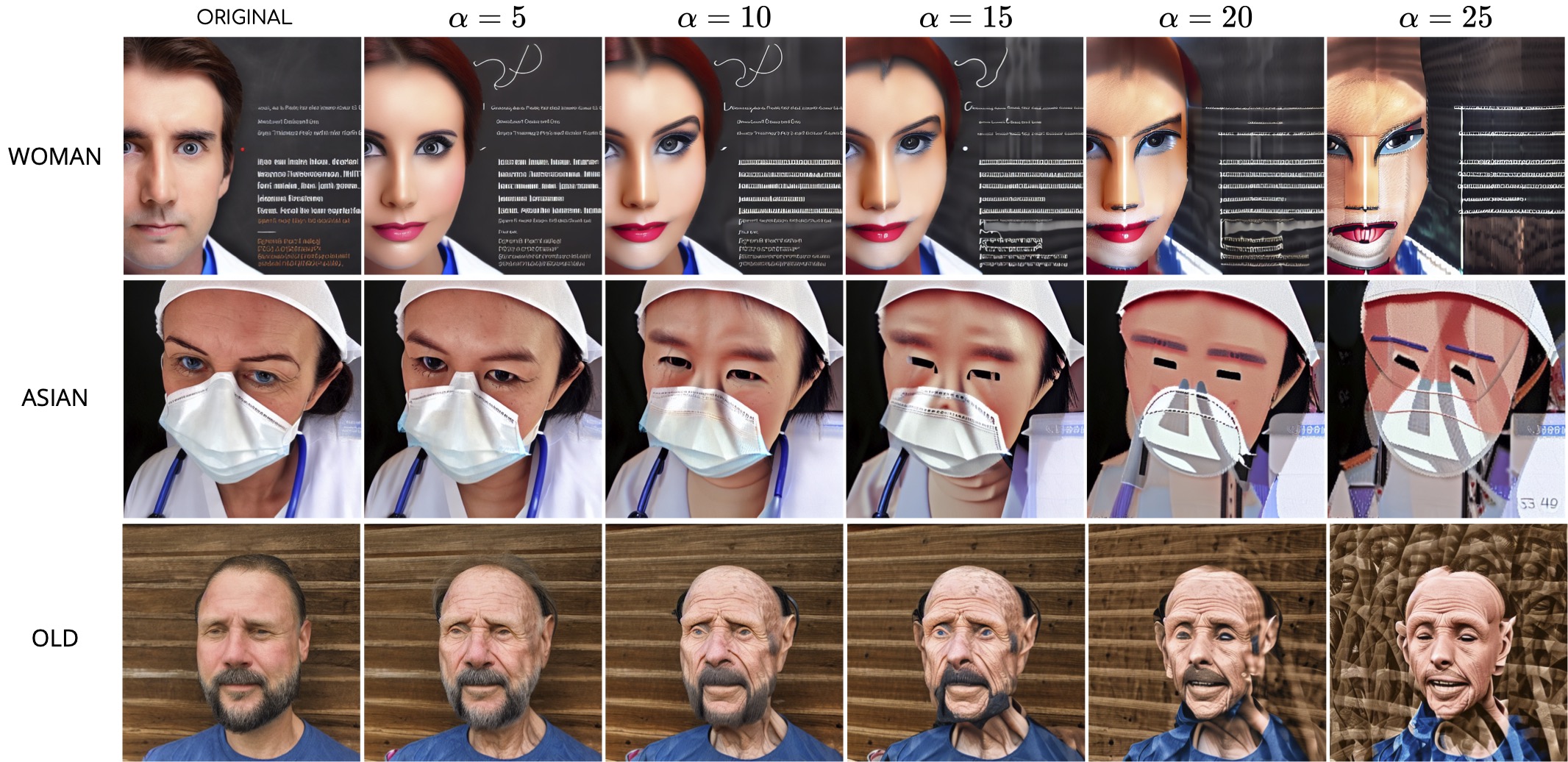}
    \caption{Example generations showing how increasing the steering strength $\alpha$ can progressively degrade the output.}
    \label{fig:appendix_alpha}
\end{figure*}

We compared several $\alpha$ configurations and observed that stronger values tend to improve the debiasing performance. However, as illustrated in Figure~\ref{fig:appendix_alpha}, excessively large $\alpha$ values introduce visible artifacts and may degrade the overall visual quality of the generated images.
 
\begin{figure}[H]
    \captionsetup{aboveskip=2pt, belowskip=2.5pt}
    \centering
    \includegraphics[width=\linewidth]{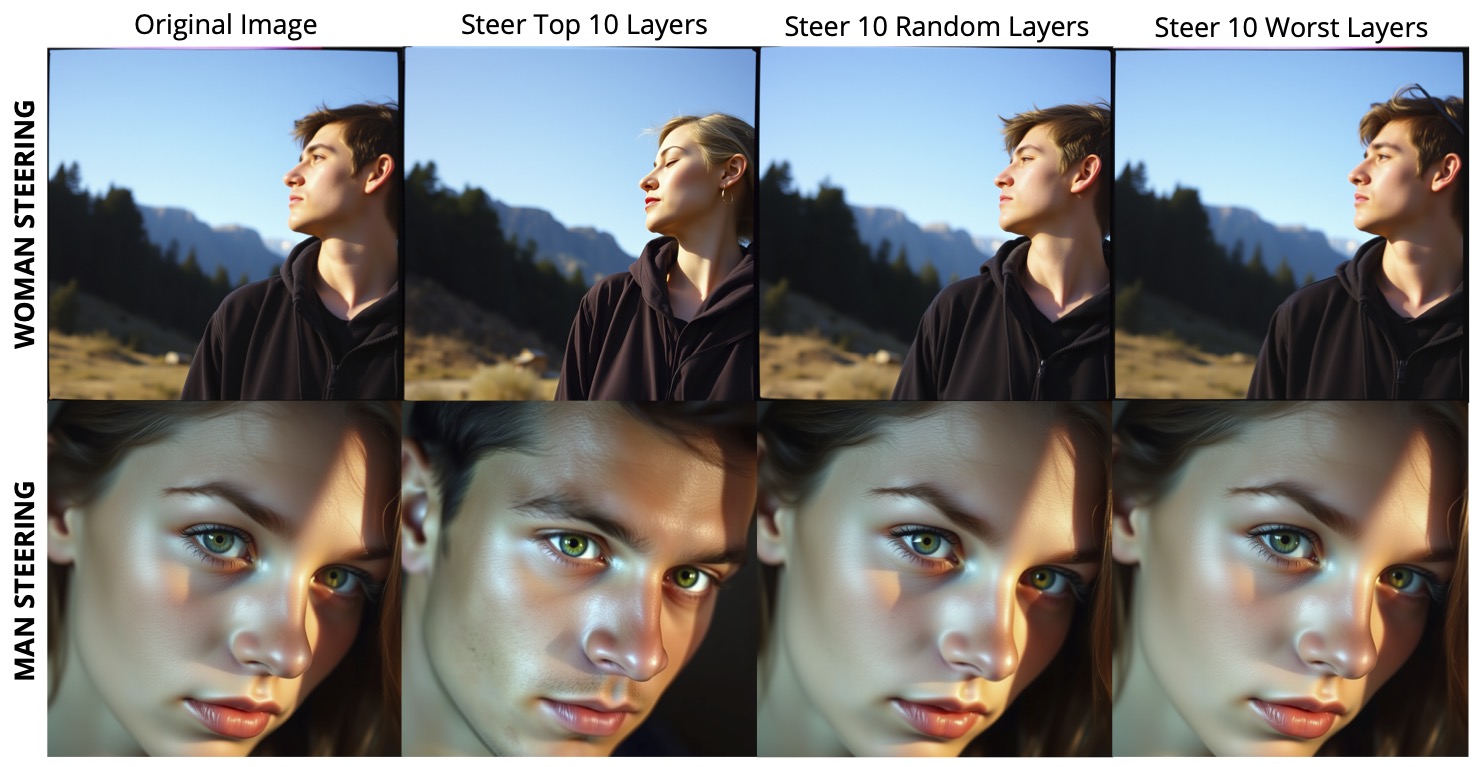}
    \caption{FLUX steering.}
    \label{fig:steering_grid_flux}
\end{figure}

 In Figure ~\ref{fig:steering_grid_flux} we present the qualitative ablation using FLUX. We evaluate both double and single stream blocks to identify the most effective layers, applying steering to the image part of each block's output. After steering 10 selected layers, the target class distribution increased from the initial $61\%$ to $98\%$ for "woman" and from $39\%$ to $95\%$ for "man", while maintaining CLIP-I scores of $0.869$ and $0.831$, respectively, demonstrating the generalization of our method to MM-DiT-based architectures.

\begin{figure}[H]
    \captionsetup{aboveskip=2pt, belowskip=2.5pt}
    \centering
    \includegraphics[width=\linewidth]{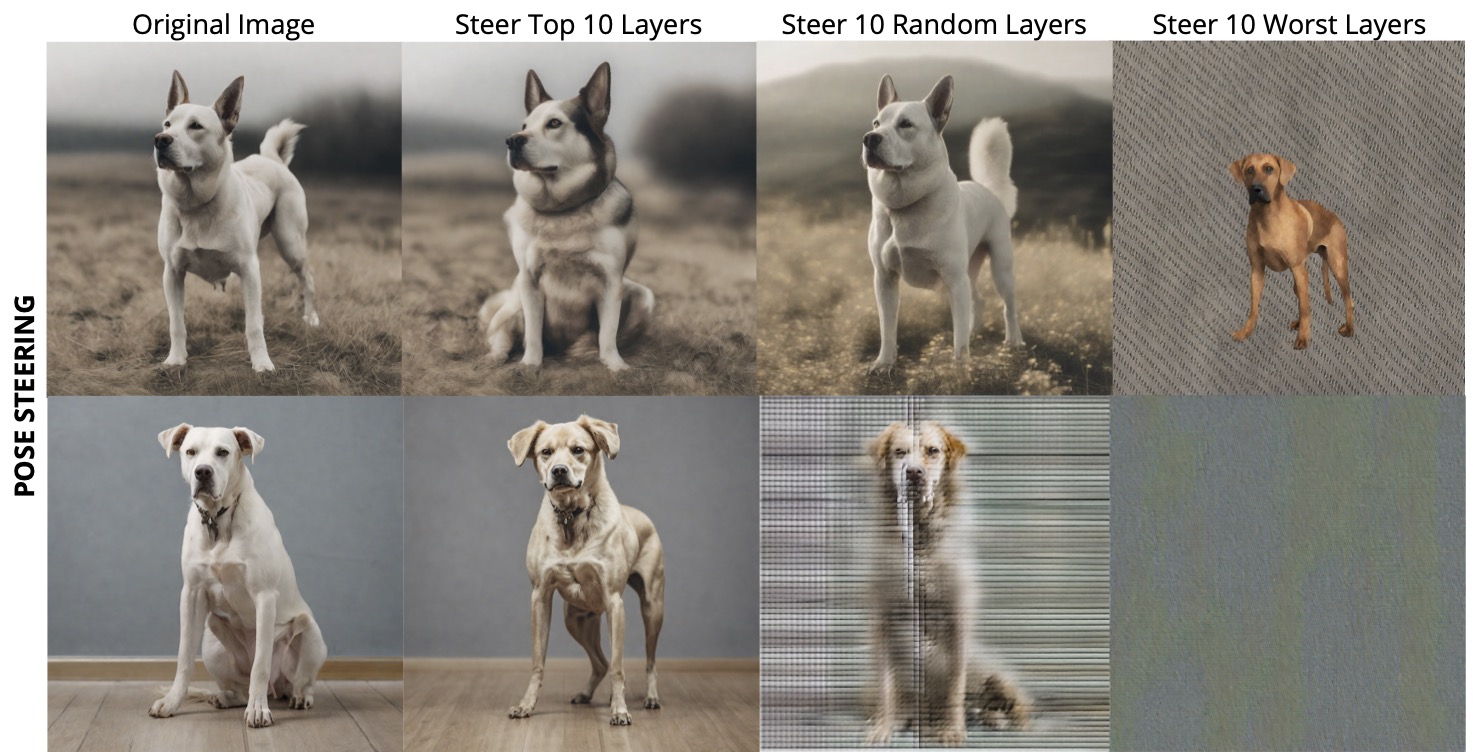}
    \caption{SDXL steering.}
    \label{fig:steering_grid_sdxl}
\end{figure}

To evaluate our method in more complex scenarios, we experiment with pose modification. Figure \ref{fig:steering_grid_sdxl} shows the effect of steering with the 10 selected layers on the dog's pose.

\section{Scalability}
\label{appendix:scalability}

While our approach involves extensive linear probing, the process is computationally efficient ($\sim$14 minutes on a 288-core CPU). We can achieve a $10\times$ speedup by utilizing a single steering vector derived from five steps; this optimized workflow yields nearly identical performance, with an $FD$ of $0.08$ and a $CLIP\text{-}I$ score of $0.89$ for gender debiasing.

We use average pooling primarily for computational feasibility, as using raw activations drastically increases the number of examples and memory requirements (e.g., for activations of shape $(1024, 640) $, $1024 \times$ more vectors). Since we calculate probes over $3$k samples, using raw activations would result in the infeasible dataset of $\approx183$~GB (for a single layer only), compared to $\approx0.2$~GB with average pooling. 
However, we run an additional experiment by sampling $10$ random patches from each activation, achieving slightly worse results compared to the average-pooled representation: FD=$0.136$ (vs $0.094$) and CLIP-I=$0.864$ (vs $0.879$).

\section{Prompt injection experiment details}
\label{appendix:prompt_injection_experiment_details}

In our experiments, we compare our localization approach with a prompt-injection-based approach, focusing on localizing social attributes related to age, gender, and race across cross-attention layers. For each target decision $D$, we construct a decision-specific dataset consisting of 
general prompts $\{p_{gen,1}, p_{gen,2}, \dots, p_{gen,N}\}$ 
(e.g., \textit{portrait of a doctor}). 
Each general prompt is paired with a collection of specific prompts 
$\{p_{spec,i,1}, p_{spec,i,2}, \dots, p_{spec,i,M}\}$ that enumerate the possible outcomes for the 
target attribute (e.g., for gender: \textit{image of a woman}). 

We provide the templates used to construct general prompts by inserting profession names:
\begin{tcolorbox}[mypromptbox]
portrait of a \{\} \\[0.4em]
face of a \{\} \\[0.4em]
a realistic portrait photo of a \{\} looking at the camera \\[0.4em]
a well-lit studio portrait of a \{\} with sharp focus \\[0.4em]
\{\} captured in a professional headshot \\[0.4em]
a \{\} at work, close-up portrait \\[0.4em]
a professional close-up headshot of a \{\} in uniform \\[0.4em]
a \{\} concentrating on their work, upper-body view \\[0.4em]
a face photograph of a \{\} with a neutral expression \\[0.4em]
a full-body photograph of a \{\} at work
\end{tcolorbox}
For each provided prompt template, we use job names associated with women: \textit{nurse}, \textit{housekeeper}, \textit{receptionist}, \textit{secretary}, and \textit{librarian}.  
Male-associated professions are \textit{construction worker}, \textit{doctor}, \textit{lawyer}, \textit{farmer}, and \textit{CEO}.

\begin{figure*}[h]
    \centering
    \includegraphics[width=0.95\textwidth]{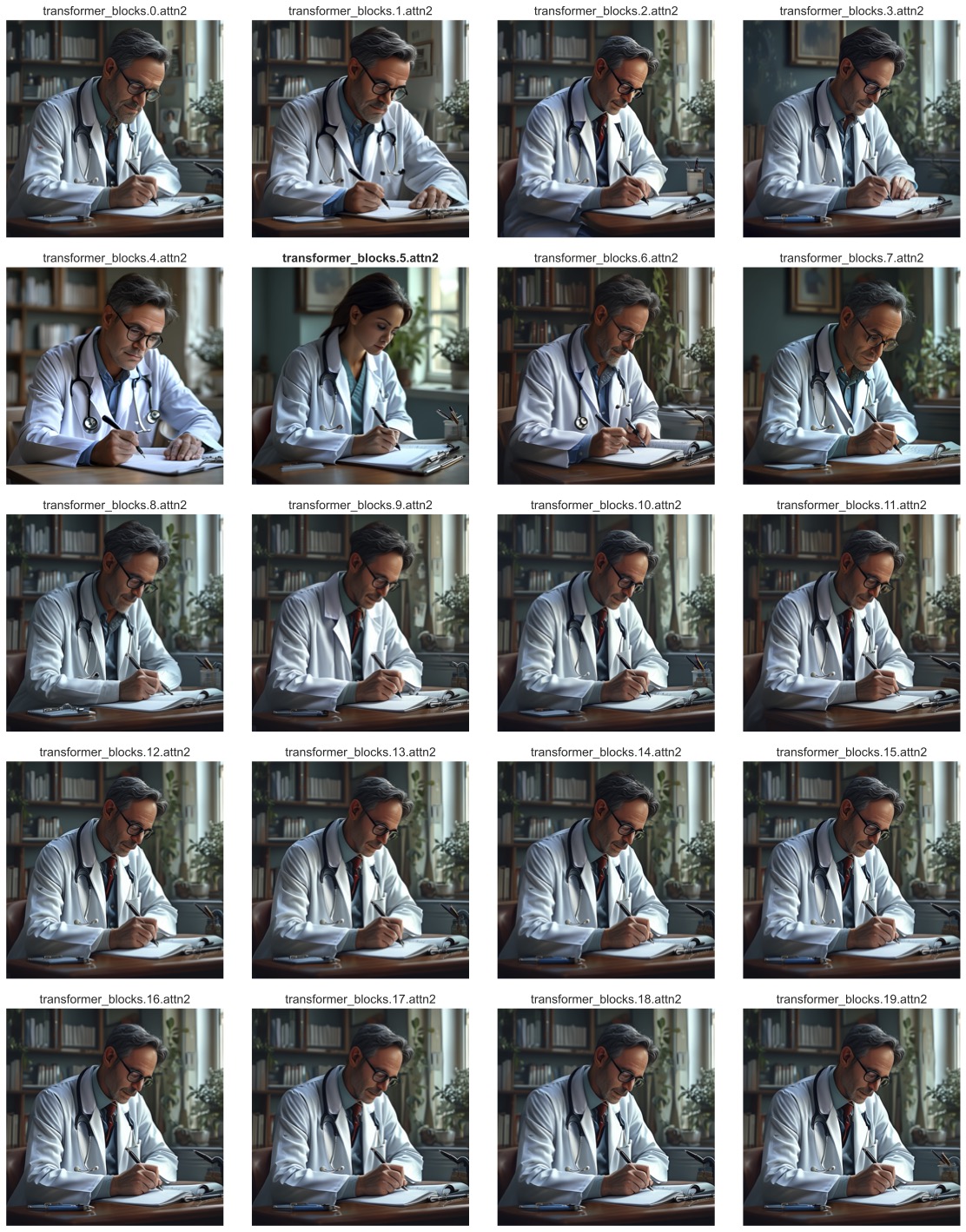}
    \caption{Example SANA images generated after injecting a specific prompt into a chosen cross-attention layer. The general prompt is \textit{“A realistic photo of a doctor sitting and writing notes”}, and the specific prompt is \textit{“a photo of a woman”}.}
    \label{fig:visualization_sana}
\end{figure*}

During each image generation, we inject a specific prompt into a single cross-attention layer across all timesteps, while all remaining layers receive the general prompt. For each prompt, we generate three images using three different seeds. We then compare the outputs with the attribute indicated by the specific prompt (e.g., \textit{image of a woman} or \textit{image of a man} when analyzing gender) to assess how the model’s decision changes. In this way, we identify the top $k$ layers that have the strongest impact on the final results. An example of prompt injection, where modifying a single layer changes the output from man to woman, is shown in Figure ~\ref{fig:visualization_sana}.

\end{document}